\documentclass[10pt,twocolumn,letterpaper]{article}

\usepackage[pagenumbers]{cvpr}

\newcommand{\ours}{MV-TAP\xspace}
\newcommand{\plucker}{\text{Pl\"{u}cker}\xspace}

\usepackage{multirow}
\usepackage{booktabs}
\usepackage[dvipsnames]{xcolor}
\usepackage{xspace}
\usepackage{colortbl}
\usepackage{pifont}
\newcommand{\cmark}{\ding{51}}
\newcommand{\xmark}{\ding{55}}
\newcommand{\paragrapht}[1]{\noindent\textbf{#1}}
\newcommand{\hlrow}{\rowcolor{black!6}}
\usepackage{algpseudocode}
\usepackage{algorithm}
\usepackage{url}
\usepackage{float}
\usepackage{wrapfig}
\usepackage[labelfont=bf]{caption}

\definecolor{cvprblue}{rgb}{0.21,0.49,0.74}
\usepackage[pagebackref,breaklinks,colorlinks,allcolors=cvprblue]{hyperref}

\def\confName{ArXiv}

\title{\ours: Tracking Any Point in Multi-View Videos}

\author{Jahyeok Koo\textsuperscript{*}\qquad
Inès Hyeonsu Kim\textsuperscript{*}\qquad
Mungyeom Kim\qquad
Junghyun Park\qquad \\
Seohyeon Park\qquad 
Jaeyeong Kim\qquad
Jung Yi\qquad
Seokju Cho\qquad 
Seungryong Kim\qquad\\[10pt]
KAIST AI\\
{\tt \href{https://cvlab-kaist.github.io/MV-TAP}{\textcolor{purple}{https://cvlab-kaist.github.io/MV-TAP}}}
}

\begin{document}
\twocolumn[{
\maketitle
\begin{center}
    \includegraphics[width=1\linewidth]{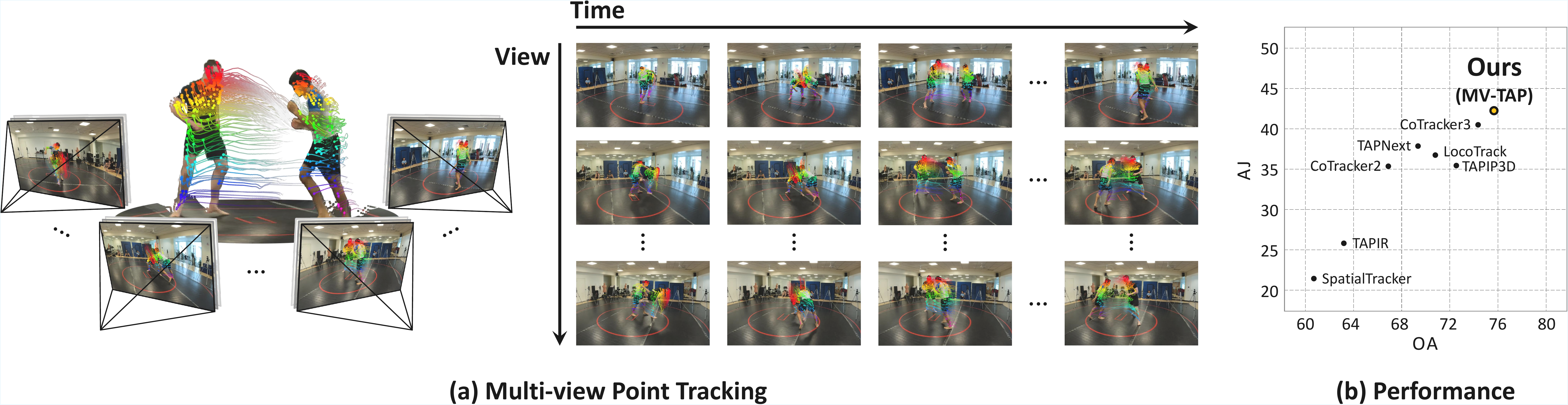}
    %\vspace{-15pt}
    \captionof{figure}{We present \textbf{\ours} (\textbf{T}racking \textbf{A}ny \textbf{P}oint in \textbf{M}ulti-view \textbf{V}ideos), a model designed to effectively integrate information across multiple viewpoints for robust and high-quality point tracking. (a) We visualize the results of \ours on Harmony4D~\cite{khirodkar2024harmony4d}. (b) \ours achieves noticeable gains over other baselines~\cite{doersch2023tapir, karaev2024cotracker, karaev2024cotracker3, cho2024local, zholus2025tapnext, xiao2024spatialtracker, zhang2025tapip3d}, demonstrating its ability to leverage multi-view information.}
    \label{fig:Teaser}
\end{center}
}]

\begin{abstract}
Multi-view camera systems enable rich observations of complex real-world scenes, and understanding dynamic objects in multi-view settings has become central to various applications. 
In this work, we present \ours, a novel point tracker that tracks points across multi-view videos of dynamic scenes by leveraging cross-view information.
\ours utilizes camera geometry and a cross-view attention mechanism to aggregate spatio-temporal information across views, enabling more complete and reliable trajectory estimation in multi-view videos. To support this task, we construct a large-scale synthetic training dataset and real-world evaluation sets tailored for multi-view tracking. Extensive experiments demonstrate that \ours outperforms existing point-tracking methods on challenging benchmarks, establishing an effective baseline for advancing research in multi-view point tracking.

\end{abstract}
\let\thefootnote\relax\footnotetext{$^*$Equal contribution}
\vspace{-0.3cm}
\section{Introduction}

Multi-view camera systems have advanced significantly in recent years, enabling the understanding of complex real-world scenes. Building upon these advances, understanding dynamic objects in multi-view systems has become crucial for a wide range of applications, including motion capture~\cite{joo2015panoptic, chatzitofis2020human4d, ghorbani2020movi, khirodkar2024harmony4d}, robot manipulation~\cite{zhao2023learning, james2020rlbench, seo2023multi}, and autonomous driving~\cite{li2025nugrounding, caesar2020nuscenes}.

Point tracking has been widely used to understand dynamic scenes by providing fine-grained spatio-temporal correspondences across video frames and indicating point-wise occlusions~\cite{harley2022particle, doersch2022tap,karaev2024cotracker3,cho2024local}. Hence, it supports a wide range of downstream tasks, including embodied AI~\cite{vecerik2024robotap, bharadhwaj2024track2act}, autonomous driving~\cite{balasingam2024drivetrack}, 4D reconstruction~\cite{wang2024shape, st4rtrack2025}, and video editing~\cite{geng2025motion, jeong2025track4gen}. Despite its strong capability, existing point tracking methods have only been explored in single-view videos. 
Since the 2D projection of a 3D scene inherently incurs geometric ambiguities, such as frequent occlusion, erratic motion, and depth uncertainty, single-view methods~\cite{harley2022particle, doersch2022tap} struggle with these challenges. Consequently, a direct application of these trackers independently to each viewpoint fails to leverage the multi-view cues required to construct reliable multi-view point trajectories.
\begin{figure}[t]
  \centering
  \includegraphics[width=\columnwidth]{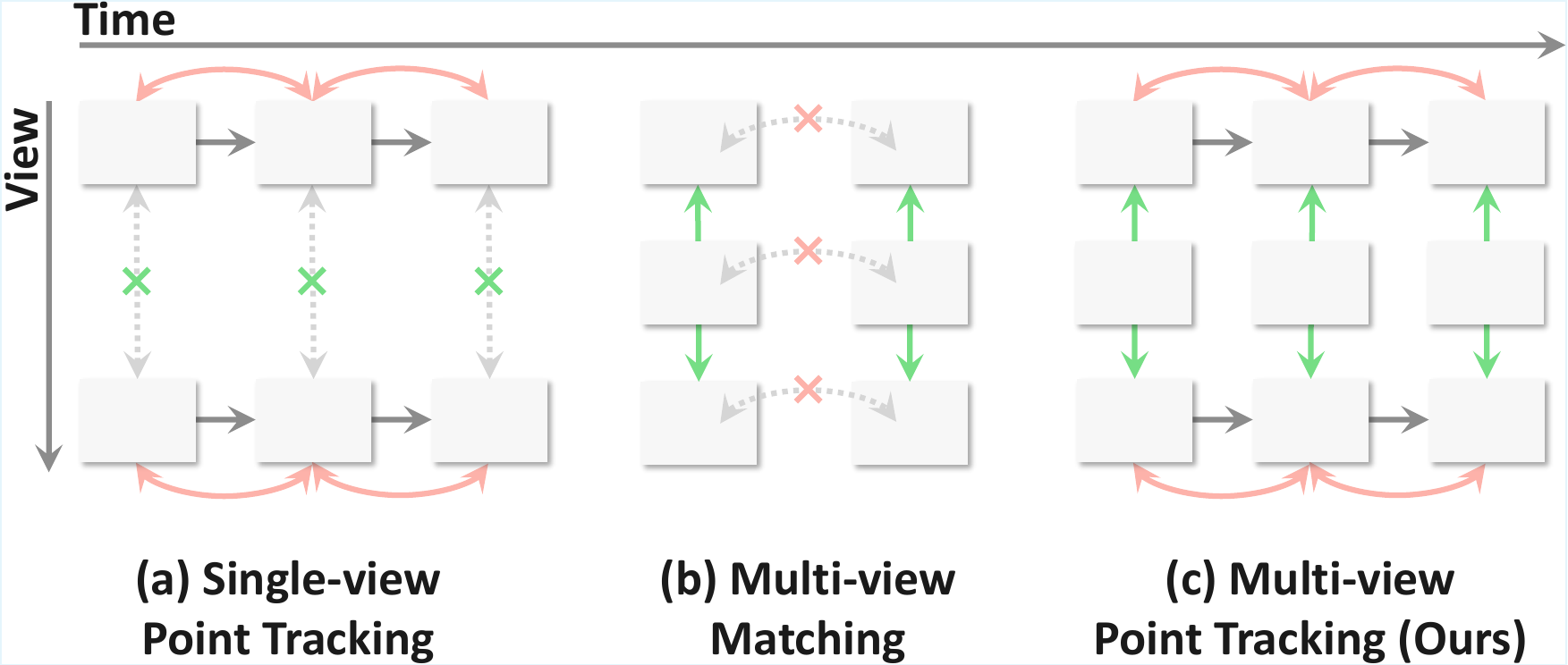}
  \caption{\textbf{Motivation.} We conceptually contrast
  our (c) multi-view point tracking with (a) single-view point tracking~\cite{doersch2023tapir, karaev2024cotracker, karaev2024cotracker3, cho2024local, zholus2025tapnext, xiao2024spatialtracker, zhang2025tapip3d} and (b) multi-view matching~\cite{lowe2004distinctive, rublee2011orb, detone2018superpoint, sarlin2020superglue, sun2021loftr}. Our approach simultaneously models both view- and frame-wise interactions to ensure cross-view and temporal consistency.}
  \label{fig:placeholder}
\end{figure}

Motivated by these observations,  we formulate the novel task of \textbf{multi-view point tracking}, which aims to track a set of query points throughout multiple videos of a dynamic scene. 
Our key intuition is that jointly processing multiple video streams of a dynamic scene provides useful spatio-temporal constraints that can help resolve the ambiguities from monocular setting. For instance, a point that is occluded or motion-blurred in one view may be clearly visible in other views. By reasoning across these viewpoints simultaneously, such ambiguities can be resolved, enabling more robust and accurate point tracking.

While multi-view correspondence has been studied as a promising way~\cite{lowe2004distinctive, rublee2011orb, detone2018superpoint, sarlin2020superglue, sun2021loftr} to capture geometric relationships, these methods are often ill-suited for dynamic point tracking. 
They typically focus on static scenes~\cite{schonberger2016structure}, assume rigid geometry, or require geometric priors~\cite{zhang2025tapip3d} that are unavailable in casual, in-the-wild videos.

Although a prior approach~\cite{rajivc2025multi} targets multi-view 3D point tracking in the 3D world coordinate system, it relies on \textit{external depth} inputs which may introduce reprojection errors when reprojecting 3D points to the 2D pixel space.
This leaves a critical gap in methodology: there is no established paradigm for leveraging only \textit{multi-view videos} to track points in pixel space through dynamic scenes.

To address these challenges, we present \textbf{\ours} (\textbf{T}racking \textbf{A}ny \textbf{P}oint in \textbf{M}ulti-view \textbf{V}ideos), which builds a holistic understanding of dynamic scenes in multi-view videos. Our framework leverages camera encoding and a cross-view attention mechanism to effectively aggregate information across all views and timesteps. Furthermore, to facilitate research in this direction, we construct a large-scale synthetic dataset specifically designed for training multi-view point-tracking models and propose an evaluation benchmark to assess generalization and robustness across diverse multi-scene settings. Experiments on multi-view benchmarks demonstrate that our approach significantly improves upon existing tracking methods, yielding more complete and accurate predictions. Our code and dataset will be made publicly available.

Our contributions are summarized as follows:
\begin{itemize}
    \item We define, for the first time, the task of multi-view point tracking in pixel space for establishing robust spatio-temporal correspondences in multi-view dynamic videos.

    \item We propose \ours, a framework that leverages camera geometry and cross-view information to address inherent limitations of single-view point tracking, such as occlusion and motion ambiguity.

    \item We demonstrate through extensive experiments that our method achieves competitive performance, providing an effective baseline for this new task.
\end{itemize}

\begin{figure*}[t]
    \centering
    \includegraphics[width=\linewidth]{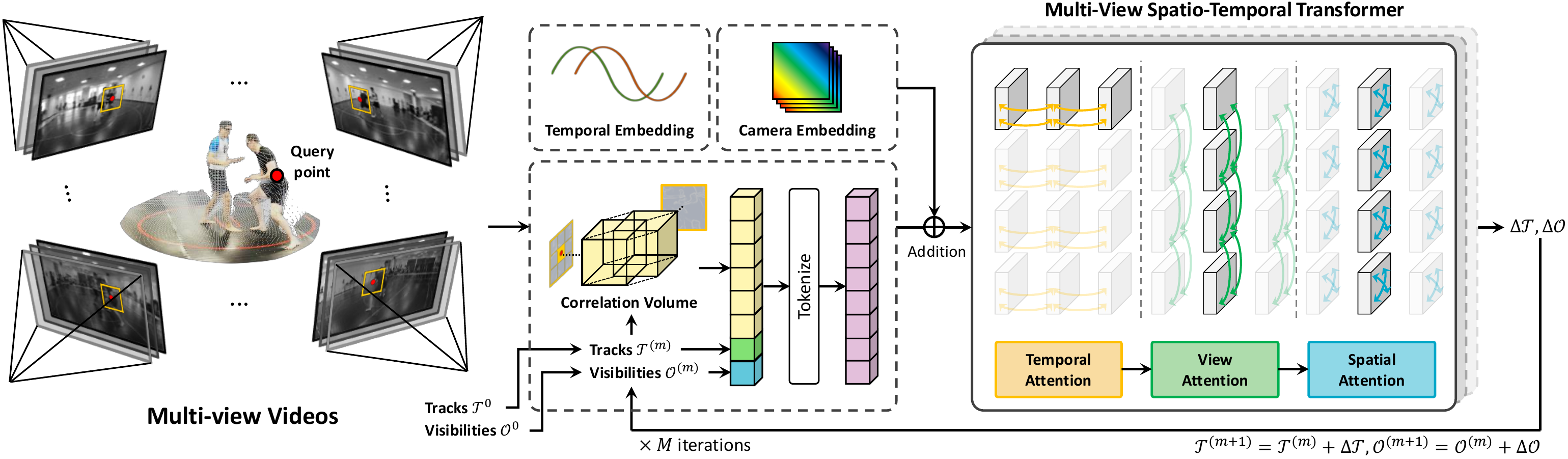}
    \vspace{-10pt}
    \caption{
        \textbf{Overall architecture of \ours.} Given synchronized multi-view videos, per-view correlation volumes are extracted from a CNN encoder feature for each query point. 
        These correlations are then tokenized and added with camera embedding for relative geometric context across views and temporal embedding. Trajectories and occlusion states are iteratively updated by a Transformer architecture, comprising temporal, spatial, and view attention modules.
    }
    \label{fig:Architecture}
    \vspace{-10pt}
\end{figure*}

\section{Related Work}

\paragrapht{Point tracking.}
The goal of point tracking is to predict the trajectories and visibilities of given query points within a video. Recent methods~\cite{harley2022particle, doersch2023tapir, xiao2024spatialtracker, koppula2024tapvid, cho2024local, karaev2024cotracker, karaev2024cotracker3, harley2025alltracker, kim2025learning, zholus2025tapnext} focus on estimating tracks by finding correspondences across frames while capturing temporal context. Specifically, PIPs~\cite{harley2022particle} proposes iterative refinement on initial query points using local correlation volumes. TAPIR~\cite{doersch2023tapir} combines this idea with a global correlation approach, which was introduced by TAP-Net~\cite{koppula2024tapvid}. LocoTrack~\cite{cho2024local} further introduces bidirectional local 4D correlation volumes for enhanced local correspondence. Notably, CoTracker~\cite{karaev2024cotracker, karaev2024cotracker3} learns the correlation between tracking points by interleaving frame- and track-wise attention modules. This approach achieves superior performance by explicitly learning the correlations between tracking points as well as temporal consistency. Unlike previous approaches, SpatialTracker~\cite{xiao2024spatialtracker} performs 3D point tracking by combining 2D tracker with depth estimation. TAPNext~\cite{zholus2025tapnext} reframes the point tracking problem as an autoregressive next-token prediction, enabling causal online tracking without specialized inductive biases. Despite these advances, existing methods are primarily designed for monocular videos, and their application to multi-view scenarios remains underexplored. 

The most relevant work to ours is MVTracker~\cite{rajivc2025multi}, which tracks query points across multi-view synchronized videos in 3D space. However, this approach highly relies on the quality of precomputed depth for lifting to 3D space. Unlike MVTracker, MV-TAP performs directly in the 2D pixel space with a view-wise attention module, showing improved robustness.

\paragrapht{Multi-view point matching.}
Multi-view point matching aims to identify correspondences for a set of query points across images captured from multiple viewpoints. Conventional methods perform pairwise matching using hand-crafted local features such as SIFT~\cite{lowe2004distinctive} and ORB~\cite{rublee2011orb}. Recent works employ learning-based approaches, such as self-supervised detector--descriptor~\cite{detone2018superpoint}, graph-based matchers~\cite{sarlin2020superglue}, and detector-free transformer matchers~\cite{sun2021loftr}. These methods can be combined with single-view trackers to extend to the multi-view setting by matching tracks across views from a reference view. However, this naive combination ignores temporal consistency within each view and is inefficient due to redundant per-frame matching.

\section{Method}
\subsection{Motivation and overview}
\label{sec:motivation}

Existing point tracking methods~\cite{cho2024local, karaev2024cotracker, karaev2024cotracker3, zholus2025tapnext, doersch2023tapir, koppula2024tapvid, zhang2025tapip3d} have mainly focused on achieving spatio-temporal consistency in single-view videos. 
While these methods excel at predicting temporally consistent trajectories, they are not designed to exploit the geometric constraints and complementary information available in a multi-view system. 

In this paper,
we define multi-view point tracking in 2D camera space.
Our goal is to leverage multi-view information to enhance tracking performance while maintaining the strong spatio-temporal consistency established by 2D trackers. To achieve this, we present \textbf{\ours} (\textbf{T}racking \textbf{A}ny \textbf{P}oint in \textbf{M}ulti-view \textbf{V}ideos), a model designed to effectively integrate information across multiple viewpoints. Our approach integrates a strong 2D tracking backbone~\cite{karaev2024cotracker3} with additional modules designed to leverage multi-view information. 
Specifically, we introduce a camera encoding module to inject geometric information and a cross view-attention module to aggregate complementary cues across viewpoints. This combination allows our model to achieve robust spatio-temporal consistency across multiple views.

First, we formulate the multi-view point tracking in pixel space (Sec.~\ref{sec:formulation}). 
We then describe the construction of our camera-aware representation (Sec.~\ref{sec:single track}), followed by multi-view spatio-temporal transformer (Sec.~\ref{sec:track head}), which effectively exploits multi-view cues through view-wise attention.
Finally, we explain our training strategy (Sec.~\ref{sec:loss}). An overview of our architecture is shown in Fig.~\ref{fig:Architecture}.

\subsection{Problem definition}
\label{sec:formulation}
The inputs of multi-view point tracking are multi-view frames $\mathcal I = \{I_{v,t} \in \mathbb R^{H\times W\times 3}\}$, query points $\mathcal Q = \{q_{v, n}\in \mathbb R^{3}\}$, and camera parameters $\mathcal{G} = \{ G_{v,t} = K [R_{v,t}| t_{v,t}]\}$ where $K\in\mathbb R^{3\times 3}$ indicates the intrinsic matrix shared across views,  $R_{v,t} \in SO(3)$ and  $t_{v,t} \in \mathbb{R}^3$ denotes rotation and translation parameters. Here, $v = 1,\dots,V$ denotes the index of camera views, $t=1,\dots,T$ refers to the frame index, and $n=1,\dots, N$ represents the index of query points.
We assume that videos from different views are temporally synchronized. 
A set of $N$ query points is independently defined for each view, since the visible time step varies across views. 
Despite being defined independently, the query points across views represent the same scene points, where each query point on view $v$ is represented by a 3-dimensional vector $q_v=(t_q, x_q, y_q)$. 
Specifically, $t_q$ denotes the queried frame index, and $(x_q, y_q)$ the spatial coordinates in that frame. 
The goal of this task is to predict a set of trajectories $\mathcal T \in \mathbb R^{V\times T\times N \times 2}$ and occlusion states $\mathcal O\in\mathbb R^{V\times T\times N\times 1}$ for the given queries, where ${\mathcal{T}}$ denotes the 2D pixel locations of $N$ points over $T$ time steps and $V$ views, and ${\mathcal{O}}$ indicates whether each point is visible or occluded across views and time.

\begin{figure*}[t]
    \centering
    \includegraphics[width=\linewidth]{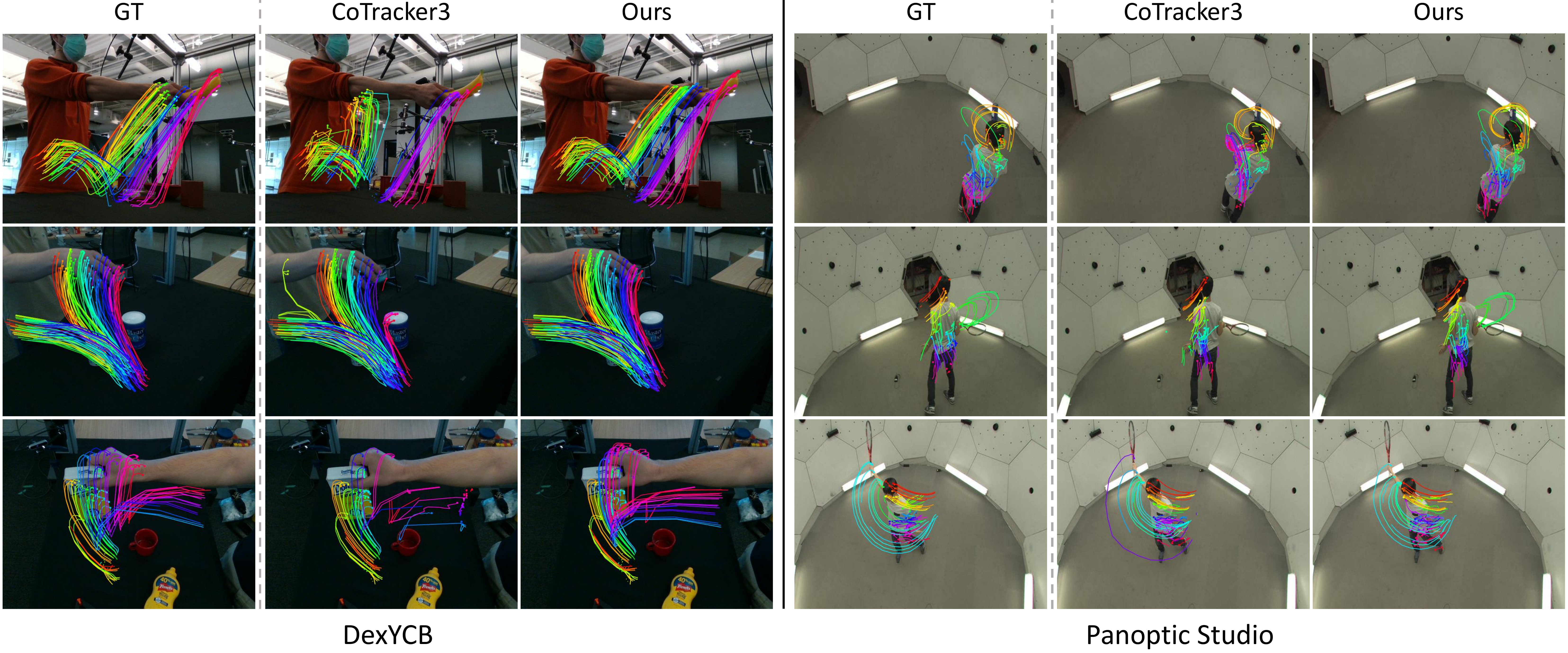}
    \vspace{-20pt}
    \caption{
        \textbf{Qualitative comparison.}
        We visualize results of \ours and a single-view baseline~\cite{karaev2024cotracker3} on the DexYCB~\cite{chao2021dexycb} and Panoptic Studio~\cite{joo2015panoptic} datasets. While the single-view baseline fails under occlusions and large motions resulting in highly fragmented tracks, \ours demonstrates superior robustness, maintaining consistent trajectories in the challenging scenarios.
        }
    \label{fig:compare_qual}
    \vspace{-5pt}
\end{figure*}

\subsection{Local 4D correlation along the temporal axis}
From initial or intermediate trajectories, we compute a cost volume as a matching representation. The intermediate trajectories are refined during this process. Following recent single-view point tracking approaches~\cite{cho2024local, karaev2024cotracker3}, we adopt local 4D correlation to exploit rich appearance cues.
Formally, given an initial or intermediate track hypothesis ${\mathcal{T}}^0$, we define the local correlation around a query point $q=(t_q, x_q, y_q)$ and its hypothesized match $p=(t_p, x_p, y_p)$.  
We construct the local neighborhoods around $p$ and $q$ within spatial radii $r_p$ and $r_q$, respectively.
The resulting local 4D correlation tensor is defined as:  
\begin{equation}
\mathcal{L}_t(i,j; p,q)
= \frac{F_t(i)\cdot F_{t_q}(j)}{\lVert F_t(i)\rVert_2\,\lVert F_{t_q}(j)\rVert_2},
\end{equation}
where $F_t$ and $F_{t_q}$ denote the feature maps from frame $t$ and the query frame $t_q$, respectively, and $i\in \mathcal{N}(p,r_p)$ and $j\in \mathcal{N}(q,r_q)$. The local 4D correlation tensor captures pairwise similarities between the two neighborhoods around $p$ and $q$.

Note that the correlation volume could also be constructed across the view dimension. However, when the baseline between viewpoints is large, the appearance similarity between corresponding local patches degrades significantly. As a result, the correlations become unreliable and may introduce noise into the model. Therefore, we compute the correlations only along the temporal dimension.

At each timestep, the correlation $\mathcal{L}_t$ and the current hypothesis position are encoded into a token. 
Stacking tokens across frames and query points forms an input token $X\in\mathbb{R}^{V\times T \times N \times d}$.

\begin{table*}[t]
  \centering
  \caption{\textbf{Quantitative comparison.}
We provide quantitative comparison of single and multi-view point trackers on DexYCB~\cite{chao2021dexycb}, Panoptic Studio~\cite{joo2015panoptic}, Kubric~\cite{greff2022kubric}, and Harmony4D~\cite{khirodkar2024harmony4d} datasets. `Target' denotes the dimension of the predicted trajectory. `Space' specifies the coordinate domain: `Camera' and `World' denotes pixel and world space, respectively. `Depth' indicates whether depth input is required. Compared to baselines, \ours achieves superior performance, demonstrating its ability to leverage multi-view information.}
  \vspace{-10pt}
  \resizebox{0.95\textwidth}{!}{
    \begin{tabular}{lccc|ccc|ccc|ccc|ccc}
    \toprule
    \multirow{2}{*}{Method} & \multirow{2}{*}{Target} & \multirow{2}{*}{Space} & \multirow{2}{*}{Depth}
    & \multicolumn{3}{c|}{\textbf{DexYCB}} 
    & \multicolumn{3}{c|}{\textbf{Panoptic Studio}}  
    & \multicolumn{3}{c|}{\textbf{Kubric}}  
    & \multicolumn{3}{c}{\textbf{Harmony4D}} \\
    & & & & AJ & $<\delta_{avg}^x$ & OA
    & AJ & $<\delta_{avg}^x$ & OA
    & AJ & $<\delta_{avg}^x$ & OA
    & AJ & $<\delta_{avg}^x$ & OA\\
    
    \midrule\midrule
    
    \multicolumn{14}{l}{\textcolor{gray}{\textit{Single-view input}}} \\
    TAPIR~\cite{doersch2023tapir} & 2D & Camera & \xmark & 29.6 & 43.9 & 66.4 & 22.1 & 39.3 & 60.0 & 66.9 & 76.9 & 89.7 & 27.6 & 53.1 & 60.0 \\
    CoTracker2~\cite{karaev2024cotracker} & 2D & Camera & \xmark & 37.5 & \textbf{62.5} & 69.4 & 33.3 & 59.1 & 64.4 & 80.5 & 91.6 & 93.7 & 37.2 & 71.9 & 55.7 \\
    LocoTrack~\cite{cho2024local} & 2D & Camera & \xmark & 38.7 & 55.8 & 74.1 & 34.9 & 56.1 & 67.5 & 82.5 & 90.1 & 93.8 & 40.8 & 72.0 & 64.1 \\
    CoTracker3~\cite{karaev2024cotracker3} & 2D & Camera & \xmark & \underline{41.5} & 59.6 & \underline{76.4} & 39.6 & 61.4 & 72.3 & 83.5 & 90.7 & 94.1 & 41.4 & 73.5 & 63.2 \\
    TAPNext~\cite{zholus2025tapnext} & 2D & Camera & \xmark & 39.6 & 57.7 & 71.9 & 36.2 & 60.1 & 66.8 & 84.6 & 93.1 & \underline{94.3} & \textbf{42.7} & \underline{74.6} & \textbf{68.4} \\
    SpatialTracker~\cite{xiao2024spatialtracker} & 3D & Camera & \cmark & 23.2 & 43.3 & 61.8 & 19.7 & 40.5 & 59.6 & 65.7 & 78.5 & 80.9 & 25.4 & 54.4 & 58.1 \\
    TAPIP3D~\cite{zhang2025tapip3d} & 3D & World & \cmark & 29.1 & 43.4 & 66.1 & \textbf{41.9} & 62.2 & \textbf{78.9} & \textbf{88.1} & \textbf{95.7} & 93.7 & 5.0 & 25.6 & 46.9 \\
    \midrule\midrule
    \multicolumn{13}{l}{\textcolor{gray}{\textit{Multi-view input}}} \\
    CoTracker3 w/ Flat. & 2D & Camera & \xmark & 2.7 & 7.1 & 35.7 & 1.0 & 12.7 & 38.8 & 19.6 & 29.3 & 34.6 & 2.1 & 20.7 & 46.4 \\
    CoTracker3 w/ Tri. & 2D & Camera & \xmark & 39.2 & 57.1 & \underline{76.4} & 37.9 & 59.5 & 72.3 & 70.2 & 82.6 & \underline{94.3} & 39.2 & 70.4 & 63.2 \\
    MVTracker~\cite{rajivc2025multi} & 3D & World & \cmark & - & 32.6 & - & - & \underline{62.4} & - & - & \underline{94.0} & - & - & 13.3 & - \\
    \hlrow \textbf{\ours} & 2D & Camera & \xmark & \textbf{44.2} & \underline{61.9} & \textbf{78.3} & \underline{40.3} & \textbf{62.8} & \underline{73.1} & \underline{87.8} & \underline{94.0} & \textbf{96.3} & \underline{42.6} & \textbf{74.9} & \underline{65.8} \\

    \bottomrule
    \end{tabular}
  }
  \label{tab:main_quan}
\end{table*}
\subsection{View-aware camera encoding}
\label{sec:single track}
In this section, we describe our view-aware camera parameter embedding. 
As camera parameters encode the 3D spatial information, we utilize them to provide the model with the relative geometric context of tracking points across views. Specifically, we leverage \plucker coordinates~\cite{plucker1828analytisch} to encode the ray corresponding to each tracking point.
Each ray $r_{v, t, n} \in \mathbb R^{6}$ is defined by \plucker coordinates:
\begin{equation}
    r=\begin{bmatrix}\mathbf {d} \\ \mathbf {m}  \end{bmatrix}, \qquad \text{where}\ \mathbf m = \mathbf{o} \times \mathbf d,
\end{equation}
where $\mathbf d \in \mathbb R^3$ is direction of the ray and $\mathbf m \in \mathbb R^3$ denotes the moment vector computed as the cross product between ray direction $\mathbf d$ and ray origin $\mathbf o \in \mathbb R^3$.
The ray direction $\mathbf d$ and ray origin $\mathbf o$ for a specific pixel are computed as follows:
\begin{equation}
    \mathbf d = R^\top K^{-1}\mathbf x, \qquad \mathbf o = -R^\top t,
\end{equation}
where $\mathbf x = (u, v, 1)^\top$ denotes the homogeneous pixel coordinate. The direction $\mathbf d$ is normalized to unit length to ensure scale-invariant representation.
Based on this definition, we construct the \plucker coordinates $\mathcal R \in \mathbb R^{V\times T\times N\times 6}$ for all trajectories and project them to a feature dimension through an MLP layer. 
The resulting embedding is then added to the input tokens, along with sinusoidal positional encoding~\cite{vaswani2017attention}. This camera encoding strategy provides the model with explicit awareness of multi-view camera geometry, allowing it to capture spatial correspondences across different views.

\subsection{Multi-view spatio-temporal transformer}
\label{sec:track head}

The encoded tokens are processed by a Transformer that interleaves \emph{temporal attention}, \emph{spatial attention}, and \emph{view attention}.
When applying attention across different axes, the feature dimension $d$ is always preserved, while the other axes are flattened into the batch dimension to enable attention along the selected axis.

\paragrapht{Temporal attention.}  
Temporal attention aggregates information along the \textbf{time axis $T$}.  
Formally, given query, key, and value projections $Q_T,K_T,V_T \in \mathbb{R}^{T \times d}$ for a fixed point,  
\begin{equation}
\mathtt{Attn}_{\text{temp}}(X) = \mathtt{Softmax}\!\left(\frac{Q_T K_T^\top}{\sqrt{d}}\right)V_T,
\end{equation}
which integrates evidence across the frame sequence, ensuring temporally smooth trajectory updates.

\paragrapht{Spatial attention.}  
Spatial attention aggregates information along the \textbf{point axis $N$} within a single frame.  
Formally, given projections $Q_N,K_N,V_N \in \mathbb{R}^{N \times d}$ for a fixed timestep,  
\begin{equation}
\mathtt{Attn}_{\text{spatial}}(X) = \mathtt{Softmax}\!\left(\frac{Q_N K_N^\top}{\sqrt{d}}\right)V_N,
\end{equation}
which helps capture rigidity priors by linking points with consistent motion patterns.

\begin{table*}[t]
    \vspace{-5pt}
    \centering
    \captionof{table}{\textbf{Ablation on various number of views}.
    We compare \ours with baselines using a varying number of input views. While performance generally improves with more views, the baselines exhibit only marginal gains. In contrast, \ours demonstrates a significantly larger improvement consistently, highlighting its superior ability to leverage multi-view information.
    }
    
    \vspace{-10pt}
    \resizebox{0.85\linewidth}{!}{
        \begin{tabular}{l|ccc|ccc|ccc|ccc}
            \toprule
            \multirow{2}{*}{Method} & \multicolumn{3}{c|}{\textbf{2 views}} & \multicolumn{3}{c|}{\textbf{4 views}} & \multicolumn{3}{c|}{\textbf{6 views}} & \multicolumn{3}{c}{\textbf{8 views}} \\
            & AJ &$<\delta^{x}_{avg}$ & OA & AJ &$<\delta^{x}_{avg}$ & OA & AJ &$<\delta^{x}_{avg}$ & OA & AJ &$<\delta^{x}_{avg}$ & OA\\
            
            \midrule\midrule
            
            CoTracker3 & \underline{37.5} & \underline{56.4} & \textbf{77.8} & \underline{38.9} & \underline{56.6} & \underline{75.1} & \underline{41.0} & \underline{58.9} & \underline{75.8} & \underline{41.5} & \underline{59.6} & \underline{76.4} \\
            CoTracker3 w/ Flat. & 9.5 & 19.5 & 54.3 & 4.4 & 8.4 & 44.5 & 3.2 & 8.3 & 39.1 & 2.7 & 7.1 & 35.7 \\
            CoTracker3 w/ Tri. & 37.1 & 55.6 & \textbf{77.8} & 37.8 & 55.3 & \underline{75.1} & 38.7 & 56.4 & \underline{75.8} & 39.2 & 57.1 & \underline{76.4} \\
            MVTracker & - & 35.8 & - & - & 31.8 & - & - & 32.8 & - & - & 32.6 & - \\
            \hlrow \textbf{\ours} & \textbf{39.2} & \textbf{56.8} & \underline{76.8} & \textbf{40.3} & \textbf{57.7} & \textbf{75.2} & \textbf{43.3} & \textbf{60.7} & \textbf{76.9} & \textbf{44.2} & \textbf{61.9} & \textbf{78.3} \\
            \bottomrule
        \end{tabular}
        \label{tab:nview}
    }
    \vspace{-10pt}
\end{table*}
\begin{table}[t]

    \centering
    \captionof{table}{\textbf{Can multi-view resolve occlusion ambiguity?} We also evaluate the position accuracy on in-frame occluded points. Our model shows robustness on the occlusion, indicating that our model effectively utilizes the multi-view cues. $<\delta^{x}_{occ}$ denotes point accuracy on in-frame occlusion points.}
    \vspace{-10pt}
    \resizebox{\linewidth}{!}{
        \begin{tabular}{l|cc|cc|cc}
            \toprule
            \multirow{2}{*}{Method} & \multicolumn{2}{c|}{\textbf{DexYCB}} & \multicolumn{2}{c|}{\textbf{Panoptic Studio}} & \multicolumn{2}{c}{\textbf{Harmony4D}}\\
            & $<\delta^{x}_{avg}$ & $<\delta^{x}_{occ}$ & $<\delta^{x}_{avg}$ & $<\delta^{x}_{occ}$ & $<\delta^{x}_{avg}$ & $<\delta^{x}_{occ}$ \\
            \midrule\midrule
            
            CoTracker3 & \underline{59.6} & 33.9 & 61.4 & 46.2 & \underline{73.5} & 58.4 \\
            CoTracker3 w/ Flat. & 7.1 & 1.9 & 12.7 & 6.3 & 20.7 & 15.0 \\
            CoTracker3 w/ Tri. & 57.1 & \underline{34.8} & 59.5 & 47.7 & 70.4 & \underline{59.1} \\
            MVTracker & 32.6 & 16.0 & \underline{62.4} & \textbf{61.2} & 13.3 & 8.9\\
            \hlrow \textbf{\ours} & \textbf{61.9} & \textbf{38.4} & \textbf{62.8} & \underline{48.7} & \textbf{74.9} & \textbf{60.3} \\
            \bottomrule
        \end{tabular}
    }
    \label{tab:occ ambiguity}
    %\vspace{-20pt}
\end{table}

\paragrapht{View attention.}  
While temporal and spatial attention capture intra-view relationships such as temporal smoothness and local rigidity, they are inherently limited in modeling inter-view relationships. 
To explicitly align representations across different views, we apply attention along the \textbf{view axis $V$}.  
Here, $Q_V,K_V,V_V \in \mathbb{R}^{V \times d}$,  
\begin{equation}
\mathtt{Attn}_{\text{view}}(X) = \mathtt{Softmax}\!\left(\frac{Q_V K_V^\top}{\sqrt{d}}\right)V_V.
\end{equation}
This attention module allows the model to exchange information between different viewpoints, thereby overcoming view-dependent ambiguities.

\subsection{Recurrent trajectory and occlusion updates}  
\label{sec:updates}
With temporal, spatial, and view attention, the Transformer iteratively refines point trajectories and occlusion probabilities. 
Concretely, at each refinement step, the Transformer predicts incremental updates to both the track position and occlusion state:
\begin{equation}
\Delta \mathcal{T}, \; \Delta \mathcal{O} = \mathtt{Transformer}(X).
\end{equation}
These updates are applied to the previous estimates as
\begin{equation}
{\mathcal{T}}^{(m+1)} = {\mathcal{T}}^{(m)} + \Delta \mathcal{T}, 
\quad 
{\mathcal{O}}^{(m+1)} = {\mathcal{O}}^{(m)} + \Delta \mathcal{O},
\end{equation}
so that after $M$ refinement steps, the model produces the final trajectory ${\mathcal{T}}$ and occlusion status ${\mathcal{O}}$.

\subsection{Training loss}
\label{sec:loss}

To train \ours, we optimize both trajectory regression and occlusion prediction.  
For trajectory supervision, we use the Huber loss~\cite{huber1992robust}:  
\begin{equation}
\mathcal{L}_{\mathrm{track}}({\mathcal{T}}, \mathcal{T}^*)
= \sum_{m=1}^{M} \gamma^{M-m}\,
\,
\ell_{\mathrm{Huber}}\!\left({\mathcal{T}}^{(m)}, \mathcal{T}^*\right),
\end{equation}
where ${\mathcal{T}}^{(m)}$ is the predicted trajectory at refinement step $m$,  
$\mathcal{T}^*$ is the ground-truth trajectory.

For occlusion supervision, we use a Binary Cross-Entropy (BCE) loss.  
We apply a sigmoid activation to the occlusion logits ${\mathcal{O}}^{(m)}$ before evaluating the loss:  
\begin{equation}
\mathcal{L}_{\mathrm{occ}}({\mathcal{O}}, \mathcal{O}^*)
= \sum_{m=1}^{M} \gamma^{M-m}\,
\mathtt{BCE}\!\left(\sigma\!\bigl({\mathcal{O}}^{(m)}\bigr), \; \mathcal{O}^*\right),
\end{equation}
where $\mathcal{O}^{(m)}$ is the predicted occlusion state at refinement step $m$ and $\mathcal{O}^*$ is the ground truth.

\section{Experiments}

\subsection{Experimental setup}
\paragrapht{Training.}
Since existing training datasets for point tracking are available only for single-view scenarios, we generate a synthetic dataset for multi-view point tracking by leveraging Kubric generation engine~\cite{greff2022kubric}.
Our generated dataset consists of synchronized multi-view videos of 5,000 scenes, along with annotations that include point trajectories, their corresponding occlusion states, and camera parameters, both intrinsic and extrinsic.
We provide further details in Sec.~\ref{sec:sup_detail}.
Our model is trained on the generated multi-view dataset for 50K steps on 4 NVIDIA A6000 GPUs with a batch size of 1 per GPU.
We employ AdamW optimizer~\cite{loshchilov2017decoupled} with a learning rate of $10^{-4}$ and a weight decay of $10^{-4}$.
We utilize a cosine learning rate scheduler with a 1,000 step warm-up stage and apply gradient clipping with a threshold of 1.0 for stable convergence.
As \ours is based on CoTracker3~\cite{karaev2024cotracker3}, we adopt the pretrained weights to initialize the feature encoder and transformer layers.
During training, we freeze only the feature extraction network, while all other parameters are updated.
The number of input views is randomly selected between 1 and 4. 
The input resolution is $384 \times 512$, and the number of trajectories is 384.
We set the number of refinement iterations to $M = 4$, and the spatial radii for local 4D correlation to $r_p=r_q=3$.

\paragrapht{Evaluation protocol.}
We evaluate our method and baselines on the DexYCB dataset~\cite{chao2021dexycb}, the Panoptic Studio dataset~\cite{joo2015panoptic}, the Kubric dataset~\cite{greff2022kubric}, and Harmony4D dataset~\cite{khirodkar2024harmony4d}. 
For DexYCB and Panoptic Studio, we utilize the point tracking annotations provided by~\cite{koppula2024tapvid} and~\cite{rajivc2025multi}, respectively, while for Kubric and Harmony4D, we construct point tracking annotations using the generation engine and human-mesh-recovery based annotation pipeline~\cite{kim2025learning}.
In the case of DexYCB, we sample dynamic points on the hand and the interacted object to focus on dynamic point tracking evaluation.
To ensure consistent evaluation, we conduct our main experiments under an 8-view setup for all datasets, where we use all 8 available views in DexYCB and Kubric, and sample 8 views from Panoptic Studio and Harmony4D. 
For view sampling, we design three sampling strategies \textit{nearest, random, and farthest} which are defined based on the distance between cameras.\begin{table}[t]
        \centering
        \captionof{table}{\textbf{Effect of additional training.} Although initialized from the same pretrained model, \ours attains consistently higher performance across all metrics. This shows that its gains primarily come from the architectural design, not merely extended training.}
         \vspace{-10pt}
        \resizebox{0.9\columnwidth}{!}{
        \begin{tabular}{l|ccc|ccc}
            \toprule
            \multirow{2}{*}{Method} & \multicolumn{3}{c|}{\textbf{DexYCB}} & \multicolumn{3}{c}{\textbf{Panoptic}} \\
            & AJ &$<\delta^{x}_{avg}$ & OA & AJ &$<\delta^{x}_{avg}$ & OA  \\
            \midrule\midrule
            CoTracker3 & \underline{41.8} & \underline{59.0} & \underline{73.8} & \underline{39.6} & \underline{61.6} & \underline{71.8} \\
            \hlrow \textbf{\ours} & \textbf{44.2} & \textbf{61.9} & \textbf{78.3} & \textbf{40.3} & \textbf{62.8} & \textbf{73.1} \\
            \bottomrule
        \end{tabular}
        }
        \label{tab:add training}
    %ㅌ\vspace{-20pt}        
\end{table}
In the main paper, we adopt the \textit{farthest} sampling strategy for our main results, while the results of other sampling strategies are provided in Sec.~\ref{sec:sup_abl}.
\begin{table*}[t]

    \begin{minipage}[t]{\textwidth}
    \centering
    \captionof{table}{\textbf{Comparison on various number of points.} We measure tracking performances under varying numbers of query points. Our model consistently outperforms shows better robustness compared to the baselines across both sparse and dense settings.}
    
    \vspace{-10pt}
    \resizebox{0.85\linewidth}{!}{
      \begin{tabular}{l|ccc|ccc|ccc|ccc}
        \toprule
        
        \multirow{2}{*}{Method} & \multicolumn{3}{c|}{\textbf{50 Points}} & \multicolumn{3}{c|}{\textbf{100 Points}} & \multicolumn{3}{c|}{\textbf{300 Points}} & \multicolumn{3}{c}{\textbf{500 Points}} \\
        & AJ &$<\delta^{x}_{avg}$ & OA & AJ &$<\delta^{x}_{avg}$ & OA & AJ &$<\delta^{x}_{avg}$ & OA & AJ &$<\delta^{x}_{avg}$ & OA \\
        \midrule\midrule
        CoTracker3 & \underline{42.0} & \underline{59.9} & \underline{74.6} & \underline{41.9} & \underline{60.1} & \underline{77.2} & \underline{41.5} & \underline{59.6} & \underline{76.4} & \underline{41.5} & \underline{59.8} & \underline{76.7} \\
        CoTracker3 w/ Flat. & 2.7 & 7.4 & 34.4 & 2.5 & 6.8 & 35.6 & 2.7 & 7.1 & 35.7 & 2.6 & 7.0 & 35.7 \\
        CoTracker3 w/ Tri. & 39.4 & 53.4 & \underline{74.6} & 39.4 & 57.1 & \underline{77.2} & 39.2 & 57.1 & \underline{76.4} & 39.5 & 57.3 & \underline{76.7} \\
        MVTracker & - & 34.2 & - & - & 32.9 & - & - & 32.6 & - & - & 34.9 & - \\
        \hlrow \textbf{\ours}  & \textbf{44.3} & \textbf{62.0} & \textbf{77.5} & \textbf{44.7} & \textbf{62.5} & \textbf{78.3} & \textbf{44.2} & \textbf{61.9} & \textbf{78.3} & \textbf{44.3} & \textbf{62.1} & \textbf{78.7} \\
        \bottomrule
      \end{tabular}
    }
    \label{tab:num point}

    \end{minipage}
    \vspace{+5pt}
    
    \begin{minipage}[t]{0.48\textwidth}
        \centering
        \captionof{table}{\textbf{Ablation on model architecture.} We present an ablation study on our model components for multi-view awareness. The performance consistently improves as each component is added, demonstrating that each module significantly contributes to leveraging multi-view information.}
        \vspace{-10pt}
        \resizebox{\textwidth}{!}{
        \begin{tabular}{l|ccc|ccc}
            \toprule
            \multirow{2}{*}{Method}
            & \multicolumn{3}{c|}{\textbf{DexYCB}} 
            & \multicolumn{3}{c}{\textbf{Panoptic Studio}}  \\
            & AJ & $<\delta_{avg}^x$  & OA
            & AJ & $<\delta_{avg}^x$  & OA  \\
            
            \midrule\midrule
            CoTracker3 & 41.5 & 59.6 & 76.4 & 39.6 & 61.4 & 72.3\\
            + View attn. & \underline{43.6} & \underline{61.5} & 77.4 & 38.6 & \underline{61.6} & 69.4 \\
            + Cam embed. & 42.2 & 60.6 & \underline{78.0} & \underline{39.9} & 60.9 & \underline{73.0} \\
            \midrule\midrule
            \hlrow \textbf{\ours} & \textbf{44.2} & \textbf{61.9} & \textbf{78.3} & \textbf{40.3} & \textbf{62.8} & \textbf{73.1} \\
        
            \bottomrule
        \end{tabular}
        }
        \label{tab:arch abl}
        %\vspace{-5pt}
    \end{minipage}
    \hfill
    \begin{minipage}[t]{\columnwidth}
        \centering
        \captionof{table}{\textbf{Comparison under frequently occluded trajectories.}  We evaluate methods on trajectories with high occlusion frequency measured by visibility-transition rate. \ours leverages cross-view cues to remain robust on frequently occluded points, improving AJ, $\delta^{x}_{\text{avg}}$ and OA.
        }
        \vspace{-10pt}
        \resizebox{\linewidth}{!}{
            \begin{tabular}{l|ccc|ccc}
                \toprule
                \multirow{2}{*}{Method} & \multicolumn{3}{c|}{\textbf{DexYCB}} & \multicolumn{3}{c}{\textbf{Panoptic Studio}} \\
                & AJ &$<\delta^{x}_{avg}$ & OA & AJ &$<\delta^{x}_{avg}$ & OA \\
                \midrule\midrule
                
                CoTracker3 & \underline{26.2} & 43.4 & \underline{66.6} & \underline{37.4} & \underline{60.6} & \underline{69.0} \\
                CoTracker3 w/ Flat. & 0.5 & 1.8 & 41.2 & 0.8 & 13.4 & 40.6 \\
                CoTracker3 w/ Tri. & 26.0 & \underline{43.6} & \underline{66.6} & 36.6 & 59.4 & \underline{69.0} \\
                MVTracker & - &  7.9 & - & - & 59.4 & - \\
                \hlrow \textbf{\ours} & \textbf{29.7} & \textbf{47.3} & \textbf{70.5} & \textbf{38.0} & \textbf{61.9} & \textbf{69.9} \\
                \bottomrule
            \end{tabular}
        }
        \label{tab:high occ}
    \end{minipage}
    %\vspace{-15pt}

\end{table*}
We use the standard point tracking metrics from TAP-Vid~\cite{doersch2022tap}, including position accuracy ($<\delta ^x_{avg}$), occlusion accuracy (OA), and Average Jaccard (AJ). 
$<\delta ^x_{avg}$ represents the average Percentage of Correct Keypoints (PCK) to evaluate the accuracy of the predicted keypoint position.
Concretely, it is computed by averaging PCK over error thresholds of 1, 2, 4, 8, and 16 pixels for visible points in ground-truth.
OA denotes the accuracy of the binary prediction for occlusion.
AJ is a composite score that jointly evaluates position and occlusion prediction of each point.

\paragrapht{Baselines.}
We compare our method against recent state-of-the-art point tracking methods, including single-view and multi-view approaches, covering both 2D and 3D formulations.
We employ TAPIR~\cite{doersch2023tapir}, CoTracker2~\cite{karaev2024cotracker}, LocoTrack~\cite{cho2024local}, CoTracker3~\cite{karaev2024cotracker3}, TAPNext~\cite{zholus2025tapnext}, SpatialTracker~\cite{xiao2024spatialtracker}, and TAPIP3D~\cite{zhang2025tapip3d} for single-view baselines and MVTracker~\cite{rajivc2025multi} for multi-view baselines. In particular, SpatialTracker, TAPIP3D and MVTracker require a depth map as input, we provide ground-truth depth maps when available. Otherwise, we utilize an off-the-shelf depth estimator~\cite{wang2024dust3r}.
Moreover, MVTracker targets 3D tracking in world space and predicts aggregated visibility across all viewpoints and thus cannot provide per-view visibility. Therefore we report only $<\delta ^x_{avg}$ for MVTracker by projecting its world-space trajectories into each camera’s pixel space.
For the single-view tracking baselines, which are designed for a monocular video setting, we perform tracking independently on each view and then aggregate the per-view results to compute the final metrics for fair comparison with multi-view methods.

Additionally, we include a naïve multi-view extension of the single-view tracker. The variant flattens the view and temporal dimensions, allowing the tracker to process multi-view videos as a single sequence.
We implement this extension with CoTracker3 as an additional multi-view baseline (\textbf{CoTracker3 w/Flat.}).
Furthermore, to examine whether simple geometric cues can resolve the monocular ambiguities, we utilize multi-view geometry with provided camera parameters. We consider two approaches, optimization with epipolar constraints or triangulation. However, the epipolar-based approach requires a reference point to define the epipolar line. Since the reference point is already noisy due to prediction inaccuracies, it leads to significant error propagation.
Consequently, we exclude this for our experiments.
Instead, we adopt triangulation-based optimization.
For each timestep, we compute a 3D point via triangulation from the 2D outputs of the single-view tracker, and then reproject it into each view to refine 2D trajectories.
The triangulation-based approach is also built upon CoTracker3 as an additional multi-view baseline (\textbf{CoTracker3 w/Tri.}).

\subsection{Main Results}
\paragrapht{Quantitative results.}
\label{sec:main quan}
We compare our approaches with recent state-of-the art point trackers on DexYCB~\cite{chao2021dexycb}, Panoptic Studio~\cite{joo2015panoptic}, Kubric~\cite{greff2022kubric} and Harmony4D~\cite{khirodkar2024harmony4d}. 
As shown in Table~\ref{tab:main_quan}, \ours achieves consistently strong performance across all benchmarks, achieving noticeable performance. While TAPIP3D~\cite{zhang2025tapip3d} and TAPNext~\cite{zholus2025tapnext} slightly outperform our methods on a few metrics, TAPIP3D relies on ground-truth depth and TAPNext adopts a substantially heavier architecture based on SSM and ViT blocks, but both methods underperform on specific benchmarks. Given that, performance of \ours indicates strong multi-view point tracking capability.

\begin{figure*}[t]
    \centering
    \includegraphics[width=\linewidth]{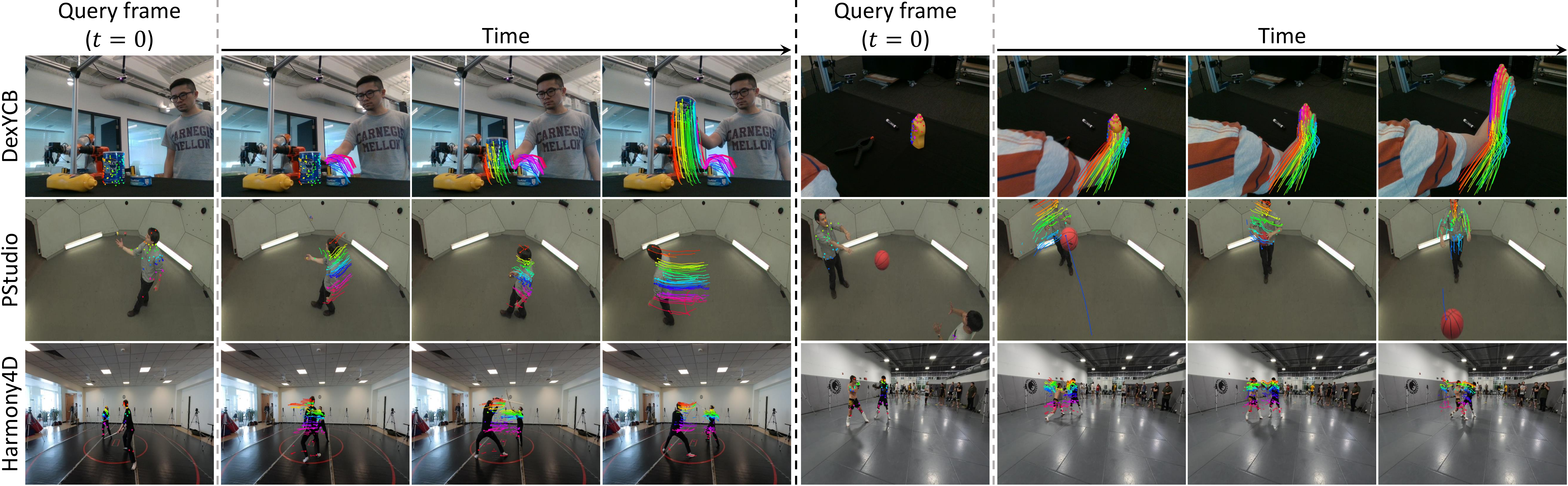}
    %\vspace{-20pt}
    \caption{\textbf{Visualization of point trajectories obtained by \ours across diverse datasets.} We showcase predictions of our model on the DexYCB~\cite{chao2021dexycb}, Panoptic Studio~\cite{joo2015panoptic}, and Harmony4D~\cite{khirodkar2024harmony4d} datasets.}
    \label{fig:main_qual}
    \vspace{-10pt}
\end{figure*}

\paragrapht{Qualitative results.}
We present qualitative comparison in Figure~\ref{fig:compare_qual} and additional qualitative results in Figure~\ref{fig:main_qual}. 
We visualize the results from the DexYCB, Panoptic Studio, and Harmony4D.
\ours shows superior robustness to large and non-rigid motions, demonstrating the effectiveness of multi-view information for point tracking.

\subsection{Ablation and Analysis}
While our main comparison includes all baselines described in Sec.~\ref{sec:main quan}, in the following ablation and analysis experiments we choose a subset of baselines for clarity and computational efficiency.

\paragrapht{Ablation on various number of views.}
Table~\ref{tab:nview} presents a comparison of \ours against a selected subset of baselines on the DexYCB\cite{chao2021dexycb} dataset, evaluating performance across various numbers of views.
Although \ours is trained only with $1$ to $4$ views due to resource limitations, owing to the attention mechanism, it can handle an arbitrary number of views even larger than $4$.
\ours consistently outperforms the baselines across various view settings and the performance of metrics steadily improves as the number of views increases. 
This indicates that additional viewpoints provide richer spatial information for multi-view tracking. 

\paragrapht{Can multi-view information resolve occlusion ambiguity?}
To investigate whether multi-view information helps resolve occlusion ambiguity, we evaluate $<\delta_\text{avg}^x$ on particular occlusion points.
In conventional point tracking evaluation pipelines, the $<\delta_\text{avg}^x$ is measured only at visible points due to the limitation of the human annotated dataset~\cite{doersch2022tap}, where occluded points are difficult to annotate.
However, automatically annotated datasets~\cite{chao2021dexycb, joo2015panoptic, greff2022kubric, khirodkar2024harmony4d} provide accurate coordinates even for occluded points that remain in the frame.
Therefore, we categorize occlusion into \textit{in-frame occlusion} and \textit{out-of-frame occlusion}, and evaluate point accuracy on in-frame occluded points.
As shown in Table~\ref{tab:occ ambiguity}, incorporating multi-view information helps with occlusion handling.
The triangulation-based optimization method shows strong robustness on in-frame occlusion points, as explicitly leveraging camera geometry helps to resolve occlusion ambiguity. However, its performance degrades on visible points due to its dependency on single-view tracker estimations.
In contrast, \ours maintains competitive stability by utilizing multi-view information, even without explicit geometric optimization.

\paragrapht{Effect of additional training.}
While \ours is initialized from the pretrained CoTracker3~\cite{karaev2024cotracker3}, we further analyze the effect of additional training. 
In Table~\ref{tab:add training}, \ours shows clear performance improvements over baseline.
This result suggests that the performance gain of \ours is not simply due to additional training, but rather benefits from its well-designed architecture which effectively leverages multi-view information to enhance generalization and robustness.

\paragrapht{Comparison on various numbers of query points.}
Downstream applications~\cite{vecerik2024robotap, bharadhwaj2024track2act, balasingam2024drivetrack, wang2024shape, st4rtrack2025, geng2025motion, jeong2025track4gen} vary in their requirements, ranging from sparse to dense correspondences estimation depending on the task.
To account for this diversity, we evaluate the tracking performance under various number of query points.
As shown in Table~\ref{tab:num point}, \ours overall achieves higher results than the baselines, highlighting its general applicability.

\paragrapht{Ablation on model components.}
In Table~\ref{tab:arch abl}, we present an ablation study on \ours architecture. We examine the effect of applying camera embedding and view attention individually and in combination. 
These results show that both modules individually improve performance over the baseline.
Notably, the combined version, which incorporates both view attention and camera embedding, achieves the best performance, demonstrating their complementary effects.

\paragrapht{Comparison on frequent occlusion scene.}
In Table~\ref{tab:high occ}, we evaluate \ours against baselines specifically on frequently occluded trajectories.
For this evaluation, we quantify the occlusion frequency, which is defined as the number of visibility state transitions per trajectory.
Then, we filter trajectories with high occlusion frequency by selecting the top 30\% with the highest occlusion frequencies.
As discussed earlier, single-view point tracking models show limited robustness in frequently occluded trajectories due to the absence of cross-view information.
In contrast, \ours maintains strong and consistent performance even under frequent occlusions, demonstrating the effectiveness of our well-designed approach in exploiting multi-view information for robust point tracking.

\section{Conclusion}
This work establishes multi-view 2D point tracking as a new and important task for advancing reliable spatio-temporal correspondence in dynamic, real-world scenes. By introducing \ours, a model that aggregates cross-view information through camera embedding and view-attention, we demonstrate how leveraging multi-view inputs can overcome key limitations of monocular trackers such as occlusion and motion ambiguity. Together with a large-scale synthetic dataset and a real-world evaluation dataset specifically designed for this task, our contributions provide both a principled formulation of the problem and a strong baseline method, paving the way for future research in robust multi-view point tracking.

\clearpage
\appendix
\section*{\Large Appendix}

%\section{Overview}
In this appendix, we first explore different strategies for constructing queries for multi-view point tracking in Sec.~\ref{supp:query_construct}. We show an illustration of the camera encoding module in Sec.~\ref{sec:camera_encoding} Next, we present additional experiments in Sec.~\ref{sec:sup_abl}. We also provide detailed descriptions of the training and evaluation datasets in Sec.~\ref{sec:sup_detail}. We then show additional qualitative results in Sec.~\ref{sec:add_qual}. Finally, we describe limitations and future work of \ours in Sec.~\ref{sec:sup_limit}.

\section{Query initialization}
\label{supp:query_construct}
We assume that the queries are given for all views.
However, this assumption is impractical for specific settings.
Thus, we explore query initialization methods by assuming that a query is given for an arbitrary view and searching for its corresponding position in the other views.
Unlike the other experiments, where queries are defined for all views as $q = (x_q, y_q, t_q)$, in this section the query is specified for a particular view. We therefore extend the definition to $q = (x_q, y_q, t_q, v_q)$ to explicitly indicate the view index.

\paragrapht{Feature-based initialization.}
First, we consider a feature matching approach, which is commonly used in conventional point tracking methods~\cite{doersch2023tapir, cho2024local} to obtain initial trajectories. We adopt ResNet pretrained on a point tracking model~\cite{karaev2024cotracker3}, VGGT~\cite{wang2025vggt}, DINOv3~\cite{simeoni2025dinov3}, and VGGT-DINO, which follows an SD-DINO~\cite{zhang2023talefeaturesstablediffusion} strategy to combine features from VGGT and DINOv3 as baselines for feature matching. Specifically, we extract VGGT features from its aggregator output.

Given a query point $q = (x_q, y_q, t_q, v_q)$, we extract its feature vector $\mathbf{f}_q$ from the backbone feature map $\Phi(V_{v_q,t_q})$ at $(x_q, y_q)$. For each candidate view $v \neq v_q$ and frame $t$, we compute a correlation map by dot product as:
\begin{equation}
C_{v,t}(x,y) = \langle \mathbf{f}_q,\ \mathbf{F}_{v,t}(x,y) \rangle,
\end{equation}
where $\mathbf{F}_{v,t} = \Phi(V_{v,t})$. We then select the most relevant frame for each view by
\begin{equation}
t_v^\ast = \arg\max_{t} \ \max_{x,y} \ C_{v,t}(x,y),
\end{equation}
and localize the 2D correspondence using argmax operation as:
\begin{equation}
(x_v^\ast, y_v^\ast) = \arg\max_{x,y} \ C_{v,t_v^\ast}(x,y).
\end{equation}
The algorithm is provided in Algorithm~\ref{alg:feature_matching}.

As shown in Tab.~\ref{tab:supp_feat_mat}, DINOv3 and VGGT-DINO show relatively strong performances compared to the other baselines, but their performance remains inadequate for query initialization.

\begin{table}[t]
    \begin{minipage}[t]{\linewidth}
        \centering
        \captionof{table}{\textbf{Comparison between feature- and depth-based initializations.} We compare query initialization using feature-based and depth-based methods in terms of $<\delta^{x}_{avg}$ on the DexYCB~\cite{chao2021dexycb} and Panoptic Studio~\cite{joo2015panoptic} datasets.}

        \vspace{-5pt}
          \resizebox{0.85\linewidth}{!}{
            \begin{tabular}{lcc}
            \toprule
            Method
            & \textbf{DexYCB} 
            & \textbf{Panoptic Studio}             
            
            \\
            
            \midrule\midrule
            
            \multicolumn{3}{l}{\textcolor{gray}{\textit{Feature-based initialization}}} \\
            ResNet & \phantom{*}\underline{7.2} & \underline{30.2} \\
            VGGT~\cite{wang2025vggt} & \phantom{*}1.5 & 17.3 \\            
            VGGT-DINO~\cite{zhang2023talefeaturesstablediffusion} & \phantom{*} 5.2 & 29.4 \\
            DINOv3~\cite{simeoni2025dinov3} & \textbf{11.8} & \textbf{36.5} \\
            \midrule
            \multicolumn{3}{l}{\textcolor{gray}{\textit{Depth-based initialization}}} \\
            MapAnything~\cite{keetha2025mapanything} & \phantom{*}\underline{6.6} & \underline{32.9} \\
            DA3~\cite{lin2025depth} & \textbf{41.6} & \textbf{47.7} \\
            \bottomrule
            \end{tabular}
            \label{tab:supp_feat_mat}
        }
      \end{minipage}       
\end{table}

\begin{table}[t]

    \vspace{-10pt}
    \begin{minipage}[t]{\linewidth}
    \centering
        \begin{algorithm}[H]
        \small
        \caption{Feature-based query initialization}
        \label{alg:feature_matching}
        
        \begin{algorithmic}[1]
        \Require Multi-view videos $V$, query point $q = (x_q, y_q, t_q, v_q)$
        \Ensure Cross-view correspondences $\{(x_v^\ast, y_v^\ast, t_v^\ast, v)\}_{v \neq v_q}$
        
        \Statex
        \State Extract query feature $\mathbf{f}_q \gets \Phi(V_{v_q,t_q})(x_q, y_q)$
        \For{each view $v \neq v_q$}
          \For{each frame $t$}
            \State $C_{v,t}(x,y) \gets \langle \mathbf{f}_q, \mathbf{F}_{v,t}(x,y) \rangle$
          \EndFor
          \State $t_v^\ast \gets \arg\max_t \max_{x,y} C_{v,t}(x,y)$
          \State $(x_v^\ast, y_v^\ast) \gets \arg\max_{x,y} C_{v,t_v^\ast}(x,y)$
        \EndFor
        \end{algorithmic}
        
        \end{algorithm}
    \end{minipage}
    \vspace{5pt}
    \begin{minipage}[t]{\linewidth}
    \centering
        \begin{algorithm}[H]
        \small
        \caption{Depth-based query initialization}
        \label{alg:depth_matching}
        
        \begin{algorithmic}[1]
        \Require Multi-view videos $V$, query point $q = (x_q, y_q, t_q, v_q)$
        \Ensure Cross-view correspondences $\{(x_v^\ast, y_v^\ast, t_v^\ast, v)\}_{v \neq v_q}$
        
        \Statex
        \State Extract depth $\mathbf{d}_{v_q, t_q} \gets \mathbf{D}(V_{v_q, t_q})(x_q, y_q)$
        \State Unproject query point to 3D
        \State $\mathbf{x}_q \gets [x_q, y_q, 1]^\top$
        \State $\mathbf{p}^{3D}_q \gets R^{\top}_{v_q,t_q} \big( d_{v_q,t_q} K^{-1} \mathbf{x}_q - t_{v_q,t_q} \big)$
        
        \State Reproject to other views
        \For{each view $v \neq v_q$}
            \State $\tilde{\mathbf{x}}_{v,t_q} \gets K \big( R_{v,t_q} \mathbf{p}^{3D}_q + t_{v,t_q} \big)$
            \State $\mathbf{x}_{v,t_q} \gets 
            \left[
                \tilde{x}_{v,t_q}/\tilde{z}_{v,t_q},\, 
                \tilde{y}_{v,t_q}/\tilde{z}_{v,t_q}
            \right]^\top$
            \State $(x^*_v, y^*_v, t^*_v) \gets (\mathbf{x}_{v,t_q},\, t_q)$
        \EndFor
        
        \end{algorithmic}
        
        \end{algorithm}
    \end{minipage}    
\end{table}

\begin{table*}[t]
  \centering
  \caption{\textbf{Threshold-wise PCK analysis on the DexYCB and Panoptic Studio datasets.}}
  \vspace{-10pt}
  \resizebox{\textwidth}{!}{
    \begin{tabular}{l|cccccc|cccccc}
    \toprule
    \multirow{2}{*}{Method} 
    & \multicolumn{6}{c|}{\textbf{DexYCB}} 
    & \multicolumn{6}{c}{\textbf{Panoptic Studio}} \\    
    & $<\delta_{1}^x$ & $<\delta_{2}^x$ & $<\delta_{4}^x$ & $<\delta_{8}^x$ & $<\delta_{16}^x$ & $<\delta_{avg}^x$
    & $<\delta_{1}^x$ & $<\delta_{2}^x$ & $<\delta_{4}^x$ & $<\delta_{8}^x$ & $<\delta_{16}^x$ & $<\delta_{avg}^x$ \\
    
    \midrule\midrule
    
    \multicolumn{13}{l}{\textcolor{gray}{\textit{Single-view input}}} \\
    TAPIR~\cite{doersch2023tapir} & 13.4 & 27.4 & 43.3 & 61.3 & 74.0 & 43.9 & 5.5 & 17.0 & 37.7 & 59.9 & 76.5 & 39.3 \\
    CoTracker2~\cite{karaev2024cotracker} & \textbf{30.8} & \textbf{44.1} & \textbf{63.2} & \textbf{82.2} & \textbf{92.0} & \textbf{62.5} & 21.9 & 40.8 & 62.4 & 79.9 & 90.5 & 59.1 \\
    LocoTrack~\cite{cho2024local} & 28.5 & 39.7 & 55.6 & 72.0 & 83.0 & 55.8 & 21.4 & 38.4 & 57.5 & 75.3 & 88.1 & 56.1 \\
    CoTracker3~\cite{karaev2024cotracker3} & 29.9 & 42.2 & 59.9 & 77.7 & 88.5 & 59.6 & \underline{23.2} & 42.6 & 65.0 & 83.0 & 93.3 & 61.4 \\
    TAPNext~\cite{zholus2025tapnext} & \underline{30.6} & 41.9 & 58.2 & 73.7 & 84.0 & 57.7 & \textbf{23.5} & 42.5 & 64.0 & 80.8 & 89.7 & 60.1 \\
    SpatialTracker~\cite{xiao2024spatialtracker} & 24.2 & 31.5 & 41.4 & 53.3 & 66.0 & 43.3 & 8.0 & 19.4 & 37.7 & 59.0 & 78.6 & 40.5 \\
    TAPIP3D~\cite{zhang2025tapip3d} & 25.2 & 36.2 & 46.0 & 52.9 & 56.7 & 43.4 & 21.2 & \textbf{43.3} & \underline{67.2} & 85.2 & 94.2 & 62.2 \\
    \midrule\midrule
    
    \multicolumn{13}{l}{\textcolor{gray}{\textit{Multi-view input}}} \\
    CoTracker3 w/ Flat. & 3.1 & 4.4 & 6.1 & 8.5 & 13.3 & 7.1 & 3.6 & 6.5 & 10.5 & 16.8 & 25.9 & 12.7 \\
    CoTracker3 w/ Tri. & 23.7 & 40.6 & 58.2 & 75.9 & 87.0 & 57.1 & 19.7 & 41.5 & 63.7 & 81.2 & 91.5 & 59.5 \\
    MVTracker~\cite{rajivc2025multi} & 19.4 & 28.5 & 35.1 & 39.0 & 41.1 & 32.6 & 15.8 & 39.7 & \textbf{69.6} & \textbf{90.0} & \textbf{96.8} & \underline{62.4} \\
    \hlrow \textbf{\ours} & 30.2 & \underline{43.5} & \underline{63.0} & \underline{81.1} & \underline{91.8} & \underline{61.9} & \underline{23.2} & \underline{42.9} & 66.3 & \underline{85.8} & \underline{95.6} & \textbf{62.8} \\

    \bottomrule
    \end{tabular}
  }
  \label{tab:supp_delta_quan_1}
\end{table*}
\begin{table*}[t]
  \centering
  \caption{\textbf{Threshold-wise PCK analysis on the Kubric and Harmony4D datasets.}}
  \vspace{-10pt}
  \resizebox{\textwidth}{!}{
    \begin{tabular}{l|cccccc|cccccc}
    \toprule
    \multirow{2}{*}{Method} 
    & \multicolumn{6}{c|}{\textbf{Kubric}} 
    & \multicolumn{6}{c}{\textbf{Harmony4D}} \\
    & $<\delta_{1}^x$ & $<\delta_{2}^x$ & $<\delta_{4}^x$ & $<\delta_{8}^x$ & $<\delta_{16}^x$ & $<\delta_{avg}^x$
    & $<\delta_{1}^x$ & $<\delta_{2}^x$ & $<\delta_{4}^x$ & $<\delta_{8}^x$ & $<\delta_{16}^x$ & $<\delta_{avg}^x$ \\
    
    \midrule\midrule
    
    \multicolumn{13}{l}{\textcolor{gray}{\textit{Single-view input}}} \\
    TAPIR~\cite{doersch2023tapir} & 43.5 & 71.4 & 84.8 & 90.8 & 94.1 & 76.9 & 11.1 & 30.2 & 56.5 & 77.7 & 90.1 & 53.1 \\
    CoTracker2~\cite{karaev2024cotracker} & 82.1 & 88.8 & 93.4 & 96.2 & 97.7 & 91.6 & 34.7 & 59.8 & 78.9 & 89.9 & 96.3 & 71.9 \\
    LocoTrack~\cite{cho2024local} & 80.2 & 87.2 & 91.7 & 94.6 & 96.7 & 90.1 & 35.2 & 59.7 & 79.0 & 90.0 & 96.1 &  72.0 \\
    CoTracker3~\cite{karaev2024cotracker3} & 81.6 & 88.2 & 92.8 & 96.0 & 98.0 & 91.3 & 34.8 & 60.8 & 81.9 & 92.4 & 97.3 & 73.5 \\
    TAPNext~\cite{zholus2025tapnext} & \textbf{85.8} & 91.2 & 94.5 & 96.4 & 97.5 & 93.1 & \textbf{36.9} & \textbf{63.0} & \underline{82.8} & \underline{92.9} & \underline{97.6} & \underline{74.6} \\
    SpatialTracker~\cite{xiao2024spatialtracker} & 59.4 & 70.8 & 80.8 & 88.1 & 93.2 & 78.4 & 15.7 & 33.8 & 57.2 & 76.0 & 89.1 & 54.4 \\
    TAPIP3D~\cite{zhang2025tapip3d} & \underline{85.5} & \textbf{95.8} & \underline{98.4} & \underline{99.4} & \underline{99.7} & \textbf{95.7} & 3.7 & 8.5 & 17.6 & 35.4 & 63.0 & 25.6 \\
    \midrule\midrule
    
    \multicolumn{13}{l}{\textcolor{gray}{\textit{Multi-view input}}} \\
    CoTracker3 w/ Flat. & 21.2 & 25.4 & 29.0 & 32.9 & 38.2 & 29.3 & 4.9 & 10.3 & 18.5 & 28.3 & 41.3 & 20.7 \\
    CoTracker3 w/ Tri. & 68.7 & 87.3 & 93.0 & 96.1 & 98.1 & 88.7 & 29.5 & 57.6 & 78.9 & 90.0 & 95.9 & 70.4 \\
    MVTracker~\cite{rajivc2025multi} & 72.8 & \underline{93.7} & \textbf{99.0} & \textbf{99.8} & \textbf{99.9} & 93.0 & 1.6 & 3.7 & 8.0 & 17.1 & 30.8 & 12.2 \\
    \hlrow \textbf{\ours} & 84.2 & 91.4 & 96.0 & 98.7 & 99.6 & \underline{94.0} & \underline{36.6} & \underline{62.8} & \textbf{83.2} & \textbf{93.4} & \textbf{98.2} & \textbf{74.9} \\

    \bottomrule
    \end{tabular}
  }
  \label{tab:supp_delta_quan_2}
\end{table*}
\begin{table*}[t]

    \vspace{-5pt}
    \begin{minipage}[t]{\textwidth}
        \centering
        \captionof{table}{\textbf{Effects of triangulation-based refinement}. `Tri' denotes whether triangulation is applied. `Final' applies triangulation to the final output, and `Window' applies it after each temporal window. `RANSAC' denotes whether outlier filtering is applied.
        }
        
        \vspace{-5pt}
        \resizebox{0.85\linewidth}{!}{
            \begin{tabular}{lcc|cc|cc|cc|cc}
                \toprule
                \multirow{2}{*}{Method} & \multirow{2}{*}{Tri.} & \multirow{2}{*}{RANSAC} & \multicolumn{2}{c|}{\textbf{DexYCB}} & \multicolumn{2}{c|}{\textbf{Panoptic Studio}} & \multicolumn{2}{c|}{\textbf{Kubric}} & \multicolumn{2}{c}{\textbf{Harmony4D}} \\
                & & & $<\delta^{x}_{avg}$ &$<\delta^{x}_{occ}$ & $<\delta^{x}_{avg}$ &$<\delta^{x}_{occ}$ & $<\delta^{x}_{avg}$ &$<\delta^{x}_{occ}$ & $<\delta^{x}_{avg}$ &$<\delta^{x}_{occ}$ \\
                
                \midrule\midrule
                
                CoTracker3 & \xmark & \xmark & \textbf{59.6} & 33.9 & \textbf{61.4} & 46.2 & \textbf{90.7} & 64.6 & \textbf{73.5} & 58.4 \\
                CoTracker3 & Final & \xmark & 52.1 & 32.5 & 58.4 & 47.2 & 85.8 & 66.2 & 69.7 & 58.9 \\
                CoTracker3 & Final & \cmark & 57.1 & \underline{34.8} & 59.5 & 47.7 & 88.7 & \underline{66.6} & 70.4 & 59.1 \\
                CoTracker3 & Window & \xmark & 53.8 & 33.9 & 59.5 & \underline{48.1} & 86.2 & 65.8 & 71.2 & \underline{60.4} \\
                CoTracker3 & Window & \cmark & \underline{58.5} & \textbf{35.5} & \underline{61.2} & \textbf{48.9} & \underline{88.9} & \textbf{67.4} & \underline{72.2} & \textbf{61.3} \\
                \midrule\midrule
                \textbf{\ours} & \xmark & \xmark & \textbf{61.9} & 38.4 & 62.8 & 48.7 & \textbf{94.0} & 70.4 & \underline{74.9} & 60.3 \\
                \textbf{\ours} & Final & \xmark & 60.6 & 39.7 & 61.2 & 50.6 & 90.9 & \underline{71.5} & 72.7 & 61.8 \\
                \textbf{\ours} & Final & \cmark & 60.5 & 39.4 & 61.6 & 50.2 & 91.2 & 71.2 & 73.1 & 62.3 \\                
                \textbf{\ours} & Window & \xmark & \underline{61.8} & \textbf{40.1} & \textbf{63.8} & \textbf{52.2} & 91.1 & \textbf{71.8} & 74.8 & \underline{64.4} \\
                \textbf{\ours} & Window & \cmark & 60.9 & 39.2 & \underline{63.2} & \underline{51.8} & \underline{91.4} & 71.3 & \textbf{75.1} & \textbf{64.7} \\
                \bottomrule
            \end{tabular}
            \label{tab:supp_optima}
        }
      \end{minipage}
    
    \vspace{10pt}

\end{table*}
\paragrapht{Depth-based initialization.}
\begin{figure}[t]
  \centering
  \includegraphics[width=\columnwidth]{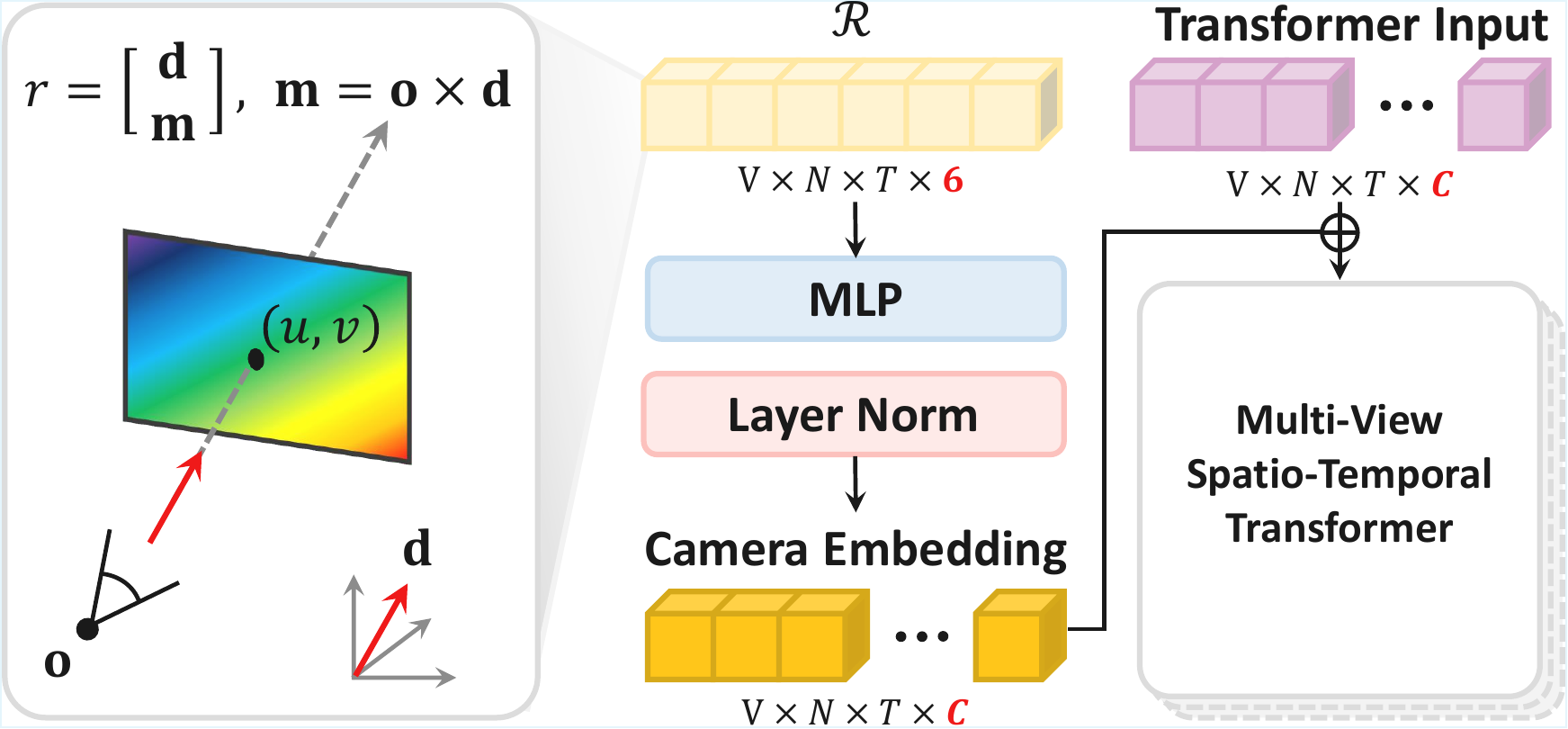}
  \caption{\textbf{Illustraction of View-aware camera encoding}. MLP-projected \plucker coordinates are added to transformer input tokens to inject view-wise information.}
  \label{fig:camera_embedding}
  \vspace{-15pt}
\end{figure}
We utilize off-the-shelf metric depth estimators, which take camera parameters as input for query initialization. 
We adopt MapAnything~\cite{keetha2025mapanything} and DepthAnything 3 (DA3)~\cite{lin2025depth} in this experiment.

First, we extract metric depth maps from the multi-view videos. 
Using the depth of the query point $\mathbf{d}_{v_q, t_q}$, we unproject the query coordinate to obtain its 3D location $\textbf{p}^{3D}_q$ using the given camera parameters:
\begin{equation}
    \mathbf{p}^{3D}_q = R^{\top}_{v_q, t_q}(\mathbf{d}_{v_q, t_q}K^{-1} \mathbf{x}_q - t_{v_q, t_q}),
\end{equation}
where $\mathbf{x}_q = [x_q, y_q, 1]^\top$ is the homogeneous coordinate of the query point, $K$ denotes the intrinsic parameters, and $R_{v_q, t_q}$ and $t_{v_q, t_q}$ are the rotation and translation parameters.
Then, we reproject the 3D coordinate of query point into the remaining views $v\neq v_q$ using their corresponding intrinsic and extrinsic parameters to obtain query initialization for multi-view point tracking.
\begin{align}
\tilde{\mathbf{x}}_{v,t_q} &= K \left( R_{v,t_q} \mathbf{p}^{3D}_q + t_{v,t_q} \right), \qquad \text{for} \ v\neq v_q, \\
\mathbf{x}_{v, t_q} &= [\tilde{x}_{v, t_q}/\tilde{z}_{v, t_q}, \tilde{y}_{v, t_q}/\tilde{z}_{v, t_q}]^\top,
\end{align}
where $\tilde{\mathbf{x}}_{v,t_q} = [\tilde{x}_{v, t_q}, \tilde{y}_{v, t_q}, \tilde{z}_{v, t_q}]^\top$ denotes the homogeneous representation of reprojected point.
The algorithm is provided in Alg.~\ref{alg:depth_matching}.

Tab.~\ref{tab:supp_feat_mat} presents the PCK of the metric depth baselines.
DepthAnything 3 (DA3) shows impressive performance, but is still insufficient for query initialization.

We analyze two types of query initialization strategies, yet none of them can properly handle the scenario where the query is provided in only a single view. Consequently, multi-view point tracking still relies on the assumption that the query is available in all views.

\section{Camera encoding}
\label{sec:camera_encoding}
In Fig.~\ref{fig:camera_embedding}, we illustrate the camera encoding introduced in Sec.~\ref{sec:single track} of the main paper. 

\section{Additional ablation and analysis}
\label{sec:sup_abl}
\begin{figure}[t]
  \centering
  \includegraphics[width=\columnwidth]{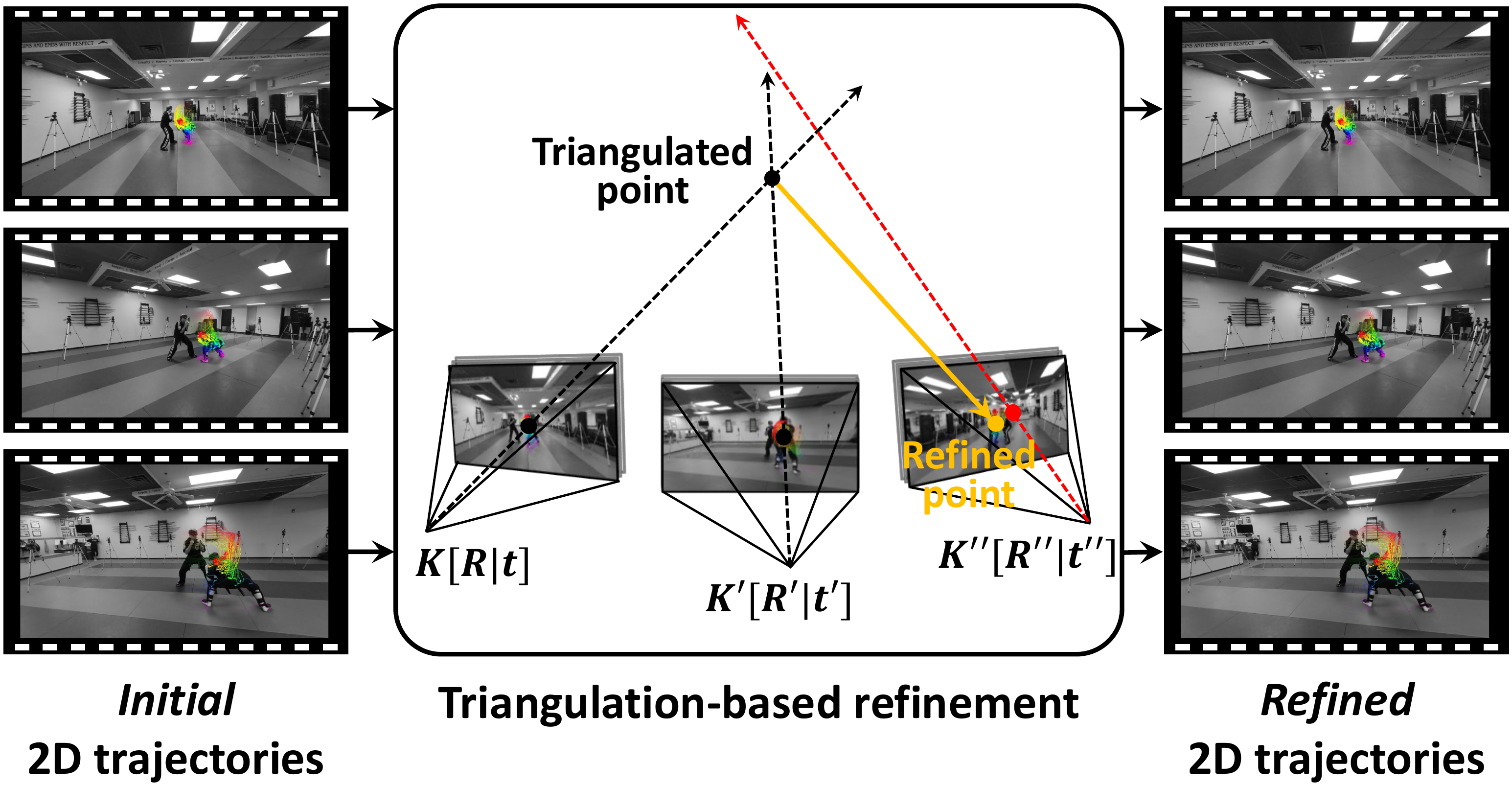}
  \vspace{-15pt}
  \caption{\textbf{Triangulation-based refinement}. For each point from initial 2D trajectories predicted by point trackers, at a given timestep, we perform triangulation with their respective camera parameters and obtain corresponding 3D point. This 3D point is then reprojected into each image to refine initial point trajectories, resolving monocular ambiguities using simple geometric cues.}
  \label{fig:suppl_tri}
  \vspace{-15pt}
\end{figure}
\paragrapht{Threshold-wise PCK analysis.}
In \cref{tab:supp_delta_quan_1,tab:supp_delta_quan_2}, we show a detailed comparison of PCK at thresholds of 1, 2, 4, 8, and 16 pixels. 
For this comparison, we use the same baselines and datasets as in the main comparison.
Across all thresholds and datasets, \ours demonstrates consistently competitive performance.

\begin{table*}[t]

    \vspace{-5pt}
    \begin{minipage}[t]{\textwidth}
        \centering
        \captionof{table}{\textbf{Occlusion Frequency}.
        We evaluate \ours and baselines on various type of frequent occlusion tracks.
        `Top-25', `Top-50', `Top-75', and `Top-100' denote the top 25\%, 50\%, 75\%, and 100\% most frequently occluded tracks, respectively.
        }
        
        \vspace{-5pt}
        \resizebox{0.85\linewidth}{!}{
            \begin{tabular}{l|ccc|ccc|ccc|ccc}
                \toprule
                \multirow{2}{*}{Method} & \multicolumn{3}{c|}{\textbf{DexYCB}} & \multicolumn{3}{c|}{\textbf{Panoptic Studio}} & \multicolumn{3}{c|}{\textbf{Kubric}} & \multicolumn{3}{c}{\textbf{Harmony4D}} \\
                & AJ &$<\delta^{x}_{avg}$ & OA & AJ &$<\delta^{x}_{avg}$ & OA & AJ &$<\delta^{x}_{avg}$ & OA & AJ &$<\delta^{x}_{avg}$ & OA\\
                
                \midrule\midrule
                \multicolumn{13}{l}{\textcolor{gray}{\textit{Top-25}}} \\
                CoTracker3 & 25.3 & 42.7 & \underline{65.9} & \underline{37.3} & \underline{60.5} & 68.9 & \underline{65.7} & 79.7 & 61.5 & \underline{33.4} & \underline{64.9} & \underline{59.2} \\
                CoTracker3 w/ Flat. & 0.5 & 1.8 & 42.3 & 0.7 & 13.2 & 40.3 & 11.9 & 23.1 & 38.2 & 1.7 & 17.4 & 49.9 \\
                CoTracker3 w/ Tri. & \underline{25.7} & \underline{43.1} & \underline{65.9} & 36.4 & 59.0 & \underline{69.1} & 65.2 & 78.8 & \underline{85.7} & 31.9 & 62.7 & \underline{59.2} \\
                MVTracker & - & 7.2 & - & - & 59.5 & - & - & \textbf{89.8} & - & - & 10.6 & - \\
                \hlrow \textbf{\ours} & \textbf{28.7} & \textbf{46.3} & \textbf{69.7} & \textbf{37.5} & \textbf{61.3} & \textbf{69.9} & \textbf{74.4} & \underline{86.6} & \textbf{90.9} & \textbf{34.6} & \textbf{67.1} & \textbf{61.7} \\
                \midrule\midrule
                \multicolumn{13}{l}{\textcolor{gray}{\textit{Top-50}}} \\
                CoTracker3 & \underline{29.3} & \underline{47.4} & \underline{68.6} & \underline{38.3} & 60.7 & \underline{70.9} & \underline{71.2} & 83.2 & 67.0 & \underline{36.1} & \underline{67.7} & \underline{60.8} \\
                CoTracker3 w/ Flat. & 0.8 & 2.8 & 40.3 & 0.7 & 12.5 & 39.7 & 0.7 & 12.5 & 39.7 & 1.9 & 18.0 & 48.4 \\
                CoTracker3 w/ Tri. & 29.0 & 46.9 & \underline{68.6} & 37.1 & 59.3 & 70.4 & 70.2 & 82.0 & \underline{88.4} & 34.3 & 65.3 & \underline{60.8} \\
                MVTracker & - & 14.1 & - & - & \underline{60.9} & - & - & \textbf{91.1} & - & - & 9.6 & - \\
                \hlrow \textbf{\ours} & \textbf{32.8} & \textbf{50.9} & \textbf{72.5} & \textbf{38.9} & \textbf{62.0} & \textbf{71.8} & \textbf{79.6} & \underline{89.5} & \textbf{93.3} & \textbf{36.8} & \textbf{69.4} & \textbf{63.0} \\
                \midrule\midrule
                \multicolumn{13}{l}{\textcolor{gray}{\textit{Top-75}}} \\
                CoTracker3 & \underline{34.2} & \underline{52.8} & \underline{72.3} & \underline{38.7} & 60.8 & \underline{71.6} & \underline{73.7} & 84.8 & \underline{89.6} & \underline{39.1} & \underline{71.0} & \underline{62.3} \\
                CoTracker3 w/ Flat. & 1.4 & 4.4 & 38.6 & 0.8 & 13.1 & 39.3 & 14.6 & 25.9 & 35.6 & 2.0 & 20.3 & 46.9 \\
                CoTracker3 w/ Tri. & 34.0 & 52.0 & \underline{72.3} & 37.7 & 59.4 & 71.5 & 72.6 & 83.5 & \underline{89.6} & 37.1 & 68.4 & \underline{62.3} \\
                MVTracker & - & 23.2 & - & - & \underline{61.4} & - & - & \textbf{91.6} & - & - & 11.5 & - \\
                \hlrow \textbf{\ours} & \textbf{37.5} & \textbf{55.7} & \textbf{74.9} & \textbf{39.6} & \textbf{62.3} & \textbf{72.6} & \textbf{81.3} & \underline{90.5} & \textbf{94.1} & \textbf{40.4} & \textbf{72.6} & \textbf{65.1} \\
                \midrule\midrule
                \multicolumn{13}{l}{\textcolor{gray}{\textit{Top-100}}} \\
                CoTracker3 & \underline{41.5} & \underline{59.6} & \underline{76.4} & \underline{39.6} & 61.4 & \underline{72.3} & \underline{83.5} & 90.7 & 94.1 & \underline{41.4} & \underline{73.5} & \underline{63.2} \\
                CoTracker3 w/ Flat. & 2.7 & 7.1 & 35.7 & 1.0 & 12.7 & 38.8 & 19.6 & 29.3 & 34.6 & 2.1 & 20.7 & 46.4 \\
                CoTracker3 w/ Tri. & 39.2 & 57.1 & \underline{76.4} & 37.9 & 59.5 & \underline{72.3} & 70.2 & 82.6 & \underline{94.3} & 39.2 & 70.4 & \underline{63.2} \\
                MVTracker & - & 32.6 & - & - & \underline{62.4} & - & - & \underline{93.0} & - & - & 12.2 & - \\
                \hlrow \textbf{\ours} & \textbf{44.2} & \textbf{61.9} & \textbf{78.3} & \textbf{40.3} & \textbf{62.8} & \textbf{73.1} & \textbf{87.8} & \textbf{94.0} & \textbf{96.3} & \textbf{42.6} & \textbf{74.9} & \textbf{65.8} \\
                \bottomrule
            \end{tabular}%
            \label{tab:supp_occ_freq}
        }
      \end{minipage}
    
\end{table*}
\begin{table*}[t]

    \vspace{-5pt}
    \begin{minipage}[t]{\textwidth}
        \centering
        \captionof{table}{\textbf{Ablation on various number of views}.
        }
        
        \vspace{-10pt}
        \resizebox{0.85\linewidth}{!}{
            \begin{tabular}{l|ccc|ccc|ccc|ccc}
                \toprule

                \multirow{2}{*}{Method} & \multicolumn{3}{c|}{\textbf{2 views}} & \multicolumn{3}{c|}{\textbf{4 views}} & \multicolumn{3}{c|}{\textbf{6 views}} & \multicolumn{3}{c}{\textbf{8 views}} \\
                & AJ &$<\delta^{x}_{avg}$ & OA & AJ &$<\delta^{x}_{avg}$ & OA & AJ &$<\delta^{x}_{avg}$ & OA & AJ &$<\delta^{x}_{avg}$ & OA\\
                \midrule\midrule
                \multicolumn{13}{l}{\textcolor{gray}{\textit{Panoptic Studio}}} \\
                CoTracker3 & \textbf{39.7} & \textbf{61.8} & \textbf{72.4} & \underline{39.4} & 60.9 & \underline{72.3} & \underline{38.6} & 59.8 & \underline{72.8} & \underline{39.6} & 61.4 & \underline{72.3} \\
                CoTracker3 w/ Flat. & 9.9 & 21.4 & 56.6 & 2.4 & 11.4 & 48.8 & 2.4 & 11.1 & 39.7 & 1.0 & 12.7 & 38.8 \\
                CoTracker3 w/ Tri. & \underline{39.6} & 61.6 & \textbf{72.4} & 38.1 & 59.5 & \underline{72.3} & 37.2 & 58.1 & \underline{72.8} & 37.9 & 59.5 & \underline{72.3} \\
                MVTracker & - & 57.0 & - & - & \underline{61.2} & - & - & \underline{60.2} & - & - & \underline{62.4} & - \\
                \hlrow \textbf{\ours} & \underline{39.6} & \underline{61.7} & \underline{72.3} & \textbf{39.8} & \textbf{61.6} & \textbf{72.5} & \textbf{39.0} & \textbf{60.7} & \textbf{73.1} & \textbf{40.3} & \textbf{62.8} & \textbf{73.1} \\
                \midrule\midrule
                \multicolumn{13}{l}{\textcolor{gray}{\textit{Harmony4D}}} \\
                CoTracker3 & 36.1 & \underline{69.8} & \underline{59.5} & \underline{38.7} & \underline{72.0} & \underline{61.3} & \underline{40.5} & \underline{72.8} & \underline{62.5} & \underline{41.4} & \underline{73.5} & \underline{63.2} \\
                CoTracker3 w/ Flat. & 14.5 & 34.1 & 51.1 & 4.3 & 20.5 & 49.8 & 3.2 & 18.2 & 47.5 & 2.1 & 20.7 & 46.4 \\
                CoTracker3 w/ Tri. & \underline{36.3} & \textbf{70.1} & \underline{59.5} & 36.8 & 69.4 & \underline{61.3} & 37.7 & 69.0 & \underline{62.5} & 39.2 & 70.4 & 63.2 \\
                MVTracker & - & 15.9 & - & - & 11.1 & - & - & 12.2 & - & - & 12.2 & - \\
                \hlrow \textbf{\ours} & \textbf{36.7} & \textbf{70.1} & \textbf{64.3} & \textbf{39.4} & \textbf{72.2} & \textbf{65.0} & \textbf{41.9} & \textbf{74.2} & \textbf{65.6} & \textbf{42.6} & \textbf{74.9} & \textbf{65.8} \\
                \bottomrule
            \end{tabular}
            \label{tab:supp_nview}
        }
      \end{minipage}
    
    \vspace{10pt}

\end{table*}
\paragrapht{Triangulation-based refinement.}
In Tab.~\ref{tab:supp_optima}, we investigate the effect of triangulation-based optimization.
We evaluate using CoTracker3~\cite{karaev2024cotracker3} and \ours, examining how different optimization strategies influence performance.
Specifically, we compare applying triangulation after each temporal window or after the entire sequence, along with the optional use of RANSAC. This process is illustrated in Fig.~\ref{fig:suppl_tri}.
We measure the performance using the average PCK of visible points and invisible in-frame points, as triangulation-based optimization directly refines the estimated 2D coordinates.

This experiment demonstrates that triangulation-based optimization tends to improve PCK ($<\delta^{x}_{occ}$ ) of the in-frame invisible points. 
While both single-view and multi-view trackers struggle to accurately estimate the positions of in-frame invisible points, triangulation-based optimization exploits reliable visible correspondences to infer more accurate estimates in these invisible regions.
However, this optimization assumes that the visible correspondences are accurate, which may lead to noisy predictions when the tracker outputs contain errors.

\begin{table}[t]

    \vspace{-5pt}
        \begin{minipage}[t]{\columnwidth}
        \centering
        \captionof{table}{\textbf{Ablation on view sampling methods}.
        }
            \resizebox{\linewidth}{!}{
            \begin{tabular}{l|ccc|ccc}
                \toprule

                \multirow{2}{*}{Method} & \multicolumn{3}{c|}{\textbf{Panoptic Studio}} & \multicolumn{3}{c}{\textbf{Harmony4D}}\\
                & AJ &$<\delta^{x}_{avg}$ & OA & AJ &$<\delta^{x}_{avg}$ & OA \\
                \midrule\midrule
                \multicolumn{7}{l}{\textcolor{gray}{\textit{Nearest Sampling}}} \\
                CoTracker3 & \underline{37.9} & \underline{58.9} & \underline{73.8} & \underline{37.9} & \underline{69.6} & \underline{61.8}\\
                CoTracker3 w/Flat. & 1.9 & 17.2 & 43.2 & 10.7 & 31.9 & 51.5\\
                CoTracker3 w/Tri. & 36.4 & 56.5 & \underline{73.8} & 36.5 & 67.5 & \underline{61.8}\\
                MVTracker & - & 57.1 & - & - & 12.0 & -\\
                \hlrow \textbf{\ours} & \textbf{39.8} & \textbf{60.3} & \textbf{75.9} & \textbf{40.4} & \textbf{71.3} & \textbf{66.5}\\
                \midrule\midrule
                \multicolumn{7}{l}{\textcolor{gray}{\textit{Random Sampling}}} \\
                CoTracker3 & \underline{40.2} & \underline{62.5} & \underline{72.7} & \underline{39.4} & \underline{72.1} & \underline{62.0}\\
                CoTracker3 w/Flat. & 1.5 & 10.5 & 49.3 & 6.5 & 20.7 & 50.5\\
                CoTracker3 w/Tri. & 38.8 & 60.7 & \underline{72.7} & 37.6 & 69.4 & \underline{62.0}\\
                MVTracker & - & 60.3 & - & - & 12.1 & -\\
                \hlrow \textbf{\ours} & \textbf{40.5} & \textbf{62.8} & \textbf{73.2} & \textbf{40.6} & \textbf{73.0} & \textbf{65.3} \\
                \bottomrule
            \end{tabular}
            \label{tab:supp_sampling}
        }
      \end{minipage}

\end{table}
\paragrapht{Comparison on frequent occlusion scene.}
We extend the frequent occlusion analysis in Tab.~\ref{tab:high occ} of main paper to further verify the generality of our approach by additionally sampling trajectories with various occlusion frequencies.
Tab.~\ref{tab:supp_occ_freq} demonstrates that \ours exhibits strong robustness to frequently occluded points.

\paragrapht{Analysis on various number of views.}
Tab.~\ref{tab:supp_nview} presents further analysis for various numbers of views on the Panoptic Studio~\cite{joo2015panoptic} and Harmony4D~\cite{khirodkar2024harmony4d} datasets.
Similar to the experiment for the DexYCB~\cite{chao2021dexycb} dataset, \ours achieves superior performance over baselines and shows steady improvements across all evaluation metrics.

\paragrapht{Analysis on sampling strategies.}
As we mentioned earlier, the Panoptic Studio~\cite{joo2015panoptic} and Harmony4D~\cite{khirodkar2024harmony4d} datasets are sampled using the \textit{farthest} strategy based on inter-camera distance.
We additionally evaluate \ours and the baselines under the \textit{nearest}, and \textit{random} sampling strategies to analyze the robustness with respect to view selection.
As shown in Tab.~\ref{tab:supp_sampling}, \ours consistently outperforms baselines across the other sampling strategies.

\section{Dataset details}
\label{sec:sup_detail}
\subsection{Training dataset}
\label{suppl:train data}
As no large-scale training dataset exists for multi-view point tracking, we introduce a new synthetic dataset generated via Kubric~\cite{greff2022kubric}. Our dataset comprises 5,000 dynamic scenes with synchronized rendered videos from four distinct viewpoints. To facilitate effective cross-view learning, camera positions are sampled in a chained manner to ensure sufficient overlap. Specifically, camera positions are sampled on hemispheres with radii ranging from 10 to 12, and adjacent viewpoints are spaced with an angular separation between $10^{\circ}$ and $45^{\circ}$. To generate globally shared ground-truth tracks, we sample query points from each viewpoint and project them onto the remaining views.

\subsection{Evaluation dataset detail}

To evaluate tracking performance on complex real-world human motion, we constructed a benchmark using the Harmony4D~\cite{khirodkar2024harmony4d} dataset, which provides synchronized multi-view videos and SMPL meshes. We obtained per-view ground-truth trajectories and visibility by projecting SMPL mesh vertices onto each image plane. To ensure the reliability of these trajectories, we employed the filtering method proposed in AnthroTAP~\cite{kim2025learning}. Following the default configuration, we filtered out erratic trajectories by comparing them with optical flow predictions~\cite{wang2024sea}, retaining only high-confidence tracks.

\section{Additional qualitative results}
\label{sec:add_qual}
\begin{figure*}[t]
    \centering
    \begin{minipage}{0.95\textwidth}
        \includegraphics[width=\linewidth]{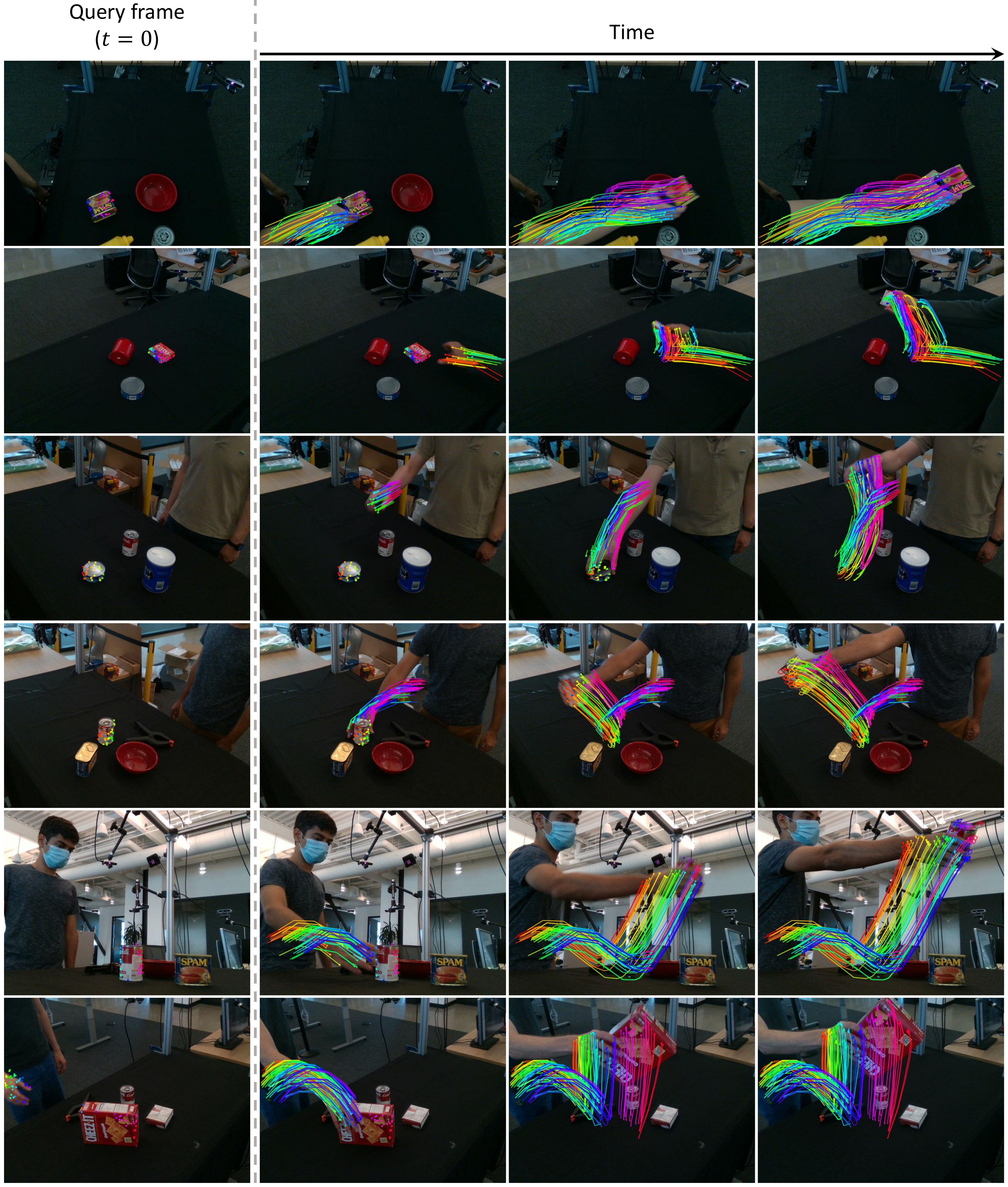}
        \vspace{-15pt}
        \captionof{figure}{\textbf{Additional qualitative results of our model.} We provide further visualizations on the DexYCB~\cite{chao2021dexycb} dataset under the 8-view evaluation setup.}
        \label{fig:suppl_qual_dex}
        \vspace{-5pt}
    \end{minipage}
\end{figure*}

\begin{figure*}[t]
    \centering
    \begin{minipage}{0.95\textwidth}
        \includegraphics[width=\linewidth]{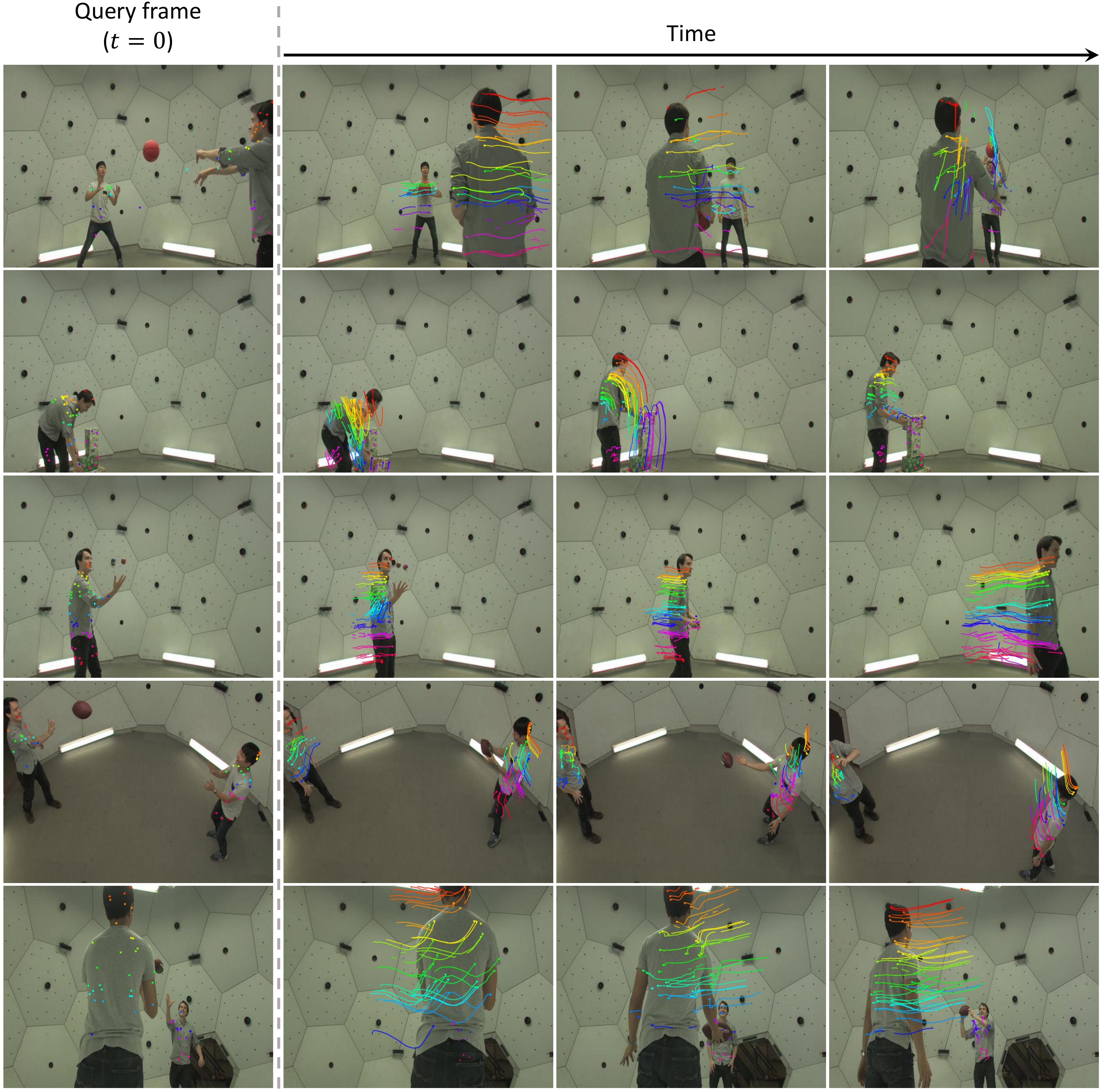}
        \vspace{-15pt}
        \captionof{figure}{\textbf{Additional qualitative results of our model.} We provide further visualizations on the Panoptic Studio~\cite{joo2015panoptic} dataset under the 8-view evaluation setup..}
        \label{fig:suppl_qual_panoptic}
        \vspace{-5pt}
    \end{minipage}
\end{figure*}

\begin{figure*}[t]
    \centering
    \begin{minipage}{0.95\textwidth}
        \includegraphics[width=\linewidth]{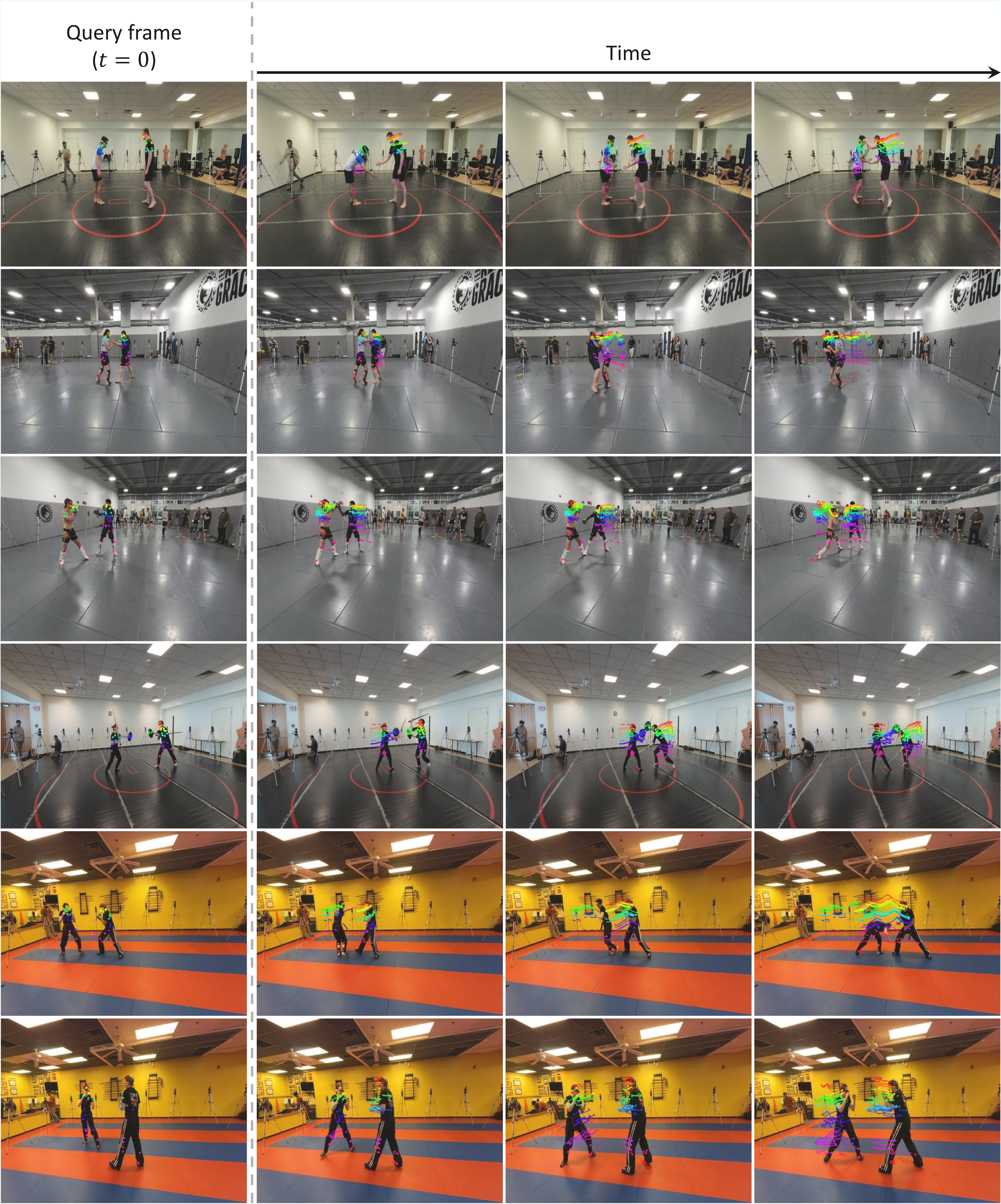}
        \vspace{-15pt}
        \captionof{figure}{\textbf{Additional qualitative results of our model.} We provide further visualizations on the Harmony4D~\cite{khirodkar2024harmony4d} dataset under the 8-view evaluation setup..}
        \label{fig:suppl_qual_harmony}
        \vspace{-5pt}
    \end{minipage}
\end{figure*}

We provide additional qualitative results in Fig.~\ref{fig:suppl_qual_dex}, Fig.~\ref{fig:suppl_qual_panoptic}, and Fig.~\ref{fig:suppl_qual_harmony}, on the DexYCB~\cite{chao2021dexycb}, Panoptic Studio~\cite{joo2015panoptic}, and Harmony4D~\cite{khirodkar2024harmony4d} datasets, respectively.

\section{Limitation and future work}
\label{sec:sup_limit}

While \ours achieves strong performance for multi-view point tracking within 2D pixel space, it still has notable limitations.
First, we assume that query points are provided in all views.
However, this requirement is often impractical. As shown in our query initialization experiments, finding correspondences for a single-view query in other views remains unreliable for robust multi-view point tracking.
Developing more advanced strategies for multi-view query initialization is therefore needed to broaden the applicability of \ours.

\clearpage

\clearpage

{
    \small
    \bibliographystyle{ieeenat_fullname}
    \bibliography{main}
}

\end{document}